\documentclass[lettersize,journal]{IEEEtran}
\usepackage{amsfonts}
\usepackage{algorithmic}
\usepackage{algorithm}
\usepackage{array}
\usepackage[caption=false,font=normalsize,labelfont=sf,textfont=sf]{subfig}
\usepackage{textcomp}
\usepackage{stfloats}
\usepackage{url}
\usepackage{verbatim}
\usepackage{graphicx}
\usepackage{cite}
\usepackage{booktabs}  % 美观的表格
\usepackage{multirow}  % 允许合并单元格
\usepackage{xcolor}  % 允许使用颜色
\usepackage{amsmath}
\usepackage{amssymb}
\begin{document}

\title{
% Towards Variable-Rate Ultra-Low Bitrate Image Compression Via Compression-Aware Continuous Rescaling Diffusion
% Compression-Aware Continuous Super-Resolution for Variable-Rate Ultra-Low Bitrate Image Compression
% Enhanced Variable-Rate Ultra-Low Bitrate Image Compression Framework Via Compression-Aware Continuous Rescaling Diffusion
% Joint Continuously Super-Resolution Enhancement Framework for Variable-Rate Ultra-Low Image Compression
% Joint Arbitrary-Scale Super-Resolution Enhancement for Variable-Rate Ultra-Low Rate Image Compression
% Towards Variable-Rate Extreme Image Compression Via Joint Degradation-Aware Arbitrary-Scale Super-Resolution Enhancement
% Towards Variable-Rate Ultra-Low Bitrate Image Compression Via Controllable Rescaling Diffusion Decoder
% Joint Degradation-Aware Arbitrary-Scale Super-Resolution Enhancement for Variable-Rate Ultra-Low Bitrate Image Compression
% Degradation-Aware Arbitrary-Scale Super-Resolution Enhanced Framework for Variable-Rate Extreme Image Compression
% Variable-Rate Extreme Image Compression Via Joint Degradation-Aware Arbitrary-Scale Super-Resolution Enhancment
Joint Degradation-Aware Arbitrary-Scale Super-Resolution for Variable-Rate Extreme Image Compression
}

\author{Xinning Chai, Zhengxue Cheng~\IEEEmembership{Member,~IEEE,}, Xin Li~\IEEEmembership{Member,~IEEE,} Rong Xie, ~\IEEEmembership{Member,~IEEE,} Li Song, ~\IEEEmembership{Senior Member,~IEEE}
        % <-this % stops a space
\thanks{Xinning Chai, Zhengxue Cheng, and Rong Xie are with the School of Information Science and Electronic Engineering, Shanghai Jiao Tong University (E-mail: chaixinning@sjtu.edu.cn, zxcheng@sjtu.edu.cn,  xierong@sjtu.edu.cn).

Xin Li is with the Department of Electronic Engineer and Information Science, University of Science and Technology of China (E-mail: xin.li@ustc.edu.cn).

Li Song is with the School of Information Science and Electronic Engineering, Shanghai Jiao Tong University and the MoE Key Lab of Artificial Intelligence, AI Institute, Shanghai Jiao Tong University, China (E-mail: song\_li@sjtu.edu.cn). \textit{(corresponding author: Zhengxue Cheng, Li Song)}}
}

\markboth{Journal of \LaTeX\ Class Files,~Vol.~14, No.~8, August~2021}%
{Shell \MakeLowercase{\textit{et al.}}: A Sample Article Using IEEEtran.cls for IEEE Journals}

\IEEEpubid{0000--0000/00\$00.00~\copyright~2021 IEEE}

\maketitle

\begin{abstract}
Recent diffusion-based extreme image compression methods have demonstrated remarkable performance at ultra-low bitrates (\textless0.1 bpp). However, most approaches require training separate diffusion models for each target bitrate, resulting in substantial computational overhead and hindering practical deployment.
Meanwhile, recent studies have shown that joint super-resolution can serve as an effective approach for enhancing low-bitrate reconstruction. However, when moving toward ultra-low-bitrate regimes, these methods struggle due to severe information loss, and their reliance on fixed super-resolution scales prevents flexible adaptation across diverse bitrates.
To address these limitations, we propose ASSR-EIC, a novel image compression framework that leverages arbitrary-scale super-resolution (ASSR) to support variable-rate extreme image compression (EIC). 
An arbitrary-scale downsampling module is introduced at the encoder side to provide controllable rate reduction, while a diffusion-based, joint degradation-aware ASSR decoder enables rate-adaptive reconstruction within a single model. We exploit the compression- and rescaling-aware diffusion prior to guide the reconstruction, yielding high-fidelity and high-realism restoration across diverse compression and rescaling settings. Specifically, we design a global compression–rescaling adaptor that offers holistic guidance for rate adaptation, and a local compression–rescaling modulator that dynamically balances generative and fidelity-oriented behaviors to achieve fine-grained, bitrate-adaptive detail restoration. To further enhance reconstruction quality, we introduce a dual semantic-enhanced design.
Extensive experiments demonstrate that ASSR-EIC delivers state-of-the-art performance in extreme image compression while simultaneously supporting flexible bitrate control and adaptive rate-dependent reconstruction.
\end{abstract}


\section{Introduction}
\IEEEPARstart{L}{ossy} image compression aims to represent images with minimal bitrates while preserving image quality. Ultra-low bitrate compression, typically operating below 0.1 bpp, is especially crucial in bandwidth-constrained scenarios, such as unmanned aerial vehicles (UAVs), aerospace imaging, and underwater exploration. However, achieving reconstructions that are simultaneously high-fidelity and high-realism under such extreme bitrate constraints remains highly challenging.

\begin{figure*}[!t]
  \centering
   \includegraphics[width=\linewidth]{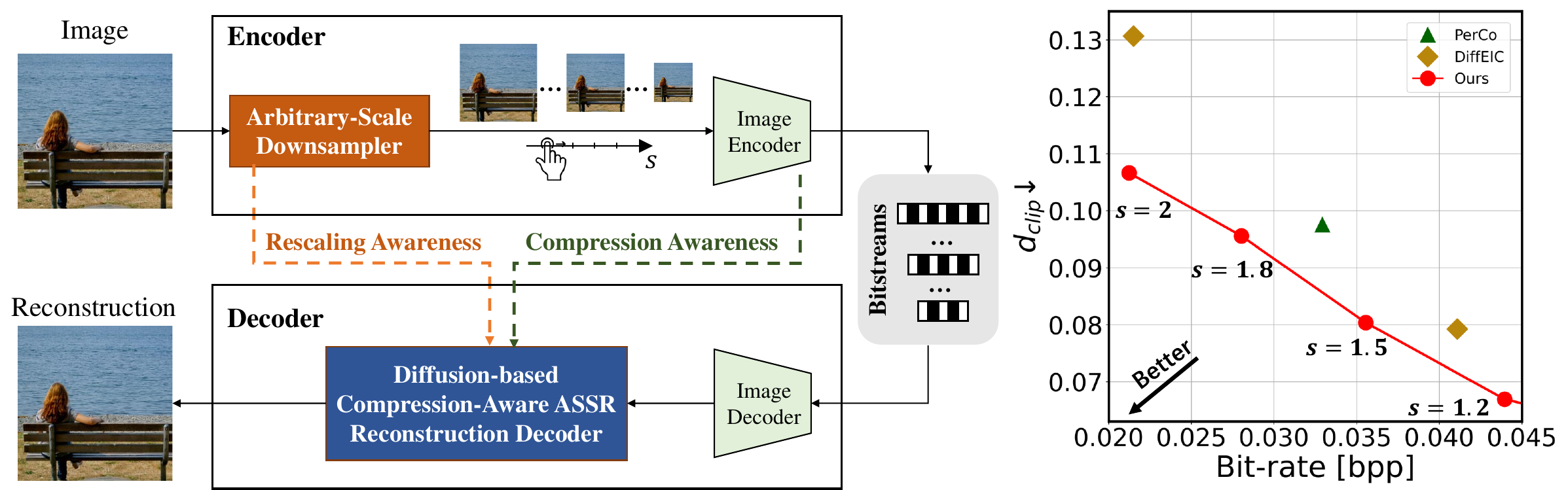}

   \caption{
   % Codec paradigm comparison (Left): (a) Existing ultra-low bitrate compression methods require retraining multiple models for different R-D points. (b) The proposed method transfers and enhances a fixed-rate model's compression capability to lower bitrates, with continuous control. $g_r$ denotes a parameter-free continuous scalable module, while $\mathcal{RD}$ refers to a plug-and-play rescaling diffusion decoder. Visual quality and bitrate comparison (Middle) with the state-of-the-art method DiffEIC \cite{li2024towards}. Our method enables continuous ultra-low bitrate (Right) compression and enhances pre-trained codecs (e.g., MS-ILLM \cite{muckley2023improving}), surpassing the state-of-the-art diffusion-based ultra-low bitrate method DiffEIC \cite{li2024towards}.
   Framework overview of the proposed ASSR-EIC and its rate–distortion performance. \textbf{Left:} Framework overview, which integrates arbitrary-scale downsampling and diffusion-based joint degradation-aware ASSR reconstruction to achieve high-quality, bitrate-adaptive extreme image compression. \textbf{Right:} R–D (CLIPScore) curves on MSCOCO demonstrate that ASSR-EIC outperforms state-of-the-art extreme image compression methods PerCo \cite{careil2024towards} and DiffEIC \cite{li2024towards}, while supporting variable bitrates, illustrated using MS-ILLM (“quality 1”) as the anchor codec with different rescaling factors $s$.
   }
   \label{fig:teaser}
   % \vspace{-1em}
\end{figure*}

Traditional compression standards such as JPEG \cite{wallace1992jpeg}, HEVC \cite{sullivan2012overview}, and VVC \cite{JVET_VTM} have been widely adopted in practice. However, when operating at low bitrates, the hand-crafted coding structure still introduce noticeable blocking artifacts. To overcome these limitations, learning-based compression approaches \cite{balle2017end,ballé2018variational,minnenjoint,cheng2020learned,he2022elic,li2023neural,lu2025learned,zhu2022transformerbased,qian2022entroformer,koyuncu2022contextformer,liu2023learned} replace the conventional components with neural networks and optimize rate–distortion performance in an end-to-end manner. While these methods achieve significant gains in objective metrics such as PSNR and MS-SSIM, their regression-based distortion objectives inherently suppress high-frequency variations and favor averaged textures. Consequently, reconstructions tend to be over-smoothed and lack realistic details, particularly at low bitrates \cite{Agustsson_2019_ICCV,mentzer2020high,muckley2023improving,agustsson2023multi,careil2024towards,li2024towards}. 

To alleviate this issue, Generative Image Compression (GIC) introduces generative modeling into the compression framework. GAN-based compression methods \cite{Agustsson_2019_ICCV,mentzer2020high,raman2020compressnet,iwai2021fidelity,agustsson2023multi,muckley2023improving} employ adversarial training to encourage perceptually realistic reconstructions. By synthesizing plausible high-frequency details instead of strictly minimizing pixel-wise distortion, these approaches improve perceptual quality in low-bitrate regimes. However, under ultra-low bitrate constraints, the encoded representation becomes highly sparse, and the limited generative capacity and training instability of GAN-based architectures make it challenging to faithfully reconstruct complex structures while preserving semantic consistency.

Recently, diffusion models have demonstrated strong generative capability and improved training stability in image synthesis tasks \cite{sohl2015deep,ho2020ddpm,songdenoising,dhariwal2021diffusion,nichol2022glide,rombach2022high}. Leveraging powerful denoising-based priors, diffusion-based compression frameworks have demonstrated strong performance in ultra-low bitrate scenarios, producing visually realistic reconstructions with rich fine-grained details \cite{theis2022lossy,ghouse2023residual,yang2023lossy,lei2023text+,li2024MISC,careil2024towards,li2024towards,kuang2024consistency,xiadiffpc}. Nevertheless, existing diffusion-based methods are typically trained for a specific bitrate setting. Accommodating diverse bandwidth conditions often requires training multiple separate models, resulting in considerable computational and deployment overhead.

\IEEEpubidadjcol
Furthermore, different bitrate regimes inherently favor different architectural paradigms. Diffusion-based generative models are particularly effective at ultra-low bitrates due to their strong prior-driven reconstruction ability. In contrast, CNN- or Transformer-based codecs optimized for distortion minimization remain advantageous at normal bitrate ranges, where sufficient signal is preserved and pixel-level fidelity is critical. This discrepancy leads to two largely distinct modeling paradigms for different bitrate segments, making it difficult to design a unified framework that performs consistently across a wide bitrate spectrum.

In light of the above observations, we seek to develop a flexible compression architecture that is compatible with both conventional and ultra-low bitrate scenarios while enabling seamless bitrate adjustment within a single framework. Along a related line of research, joint super-resolution (SR)-enhanced compression \cite{khani2021efficient,tian2021self,li2022efficient,amirpour2022deepstream,luo2022livesr,jeong2024real,zhao2025instance,yu2025stsr360} has been explored to improve coding efficiency at low bitrates by transmitting a downsampled representation at the encoder and restoring the original resolution at the decoder via SR. While this strategy demonstrates that compression and resolution reconstruction can be jointly optimized, existing approaches typically employ fixed-scale SR models (e.g., ×2 or ×4) and are designed for moderate bitrate reduction within conventional ranges. Their limited scalability and regression-based architectures restrict their applicability to ultra-low bitrate scenarios and prevent flexible bitrate control across diverse operating conditions.

Building upon this rescaling paradigm, we further incorporate arbitrary-scale super-resolution (ASSR) together with compression-aware diffusion-based reconstruction to enable continuous and controllable scaling within a unified generative framework. Specifically, we introduce arbitrary-scale downsampled encoding and scale-adaptive generative reconstruction, allowing flexible bitrate scalability under varying compression strengths and scaling factors. Moreover, the ultra-low bitrate components are implemented in a plug-and-play manner, ensuring compatibility with existing conventional codecs and reducing system modification effort and deployment cost.

Nevertheless, realizing such a unified and scalable framework presents several key challenges. First, ultra-low bitrate compression introduces severe and irreversible information loss, significantly reducing structural and textural cues and rendering reconstruction increasingly ill-posed as compression becomes more aggressive.
Second, arbitrary encoder-side downsampling reshapes the spatial distribution and amount of preserved information prior to compression, thereby directly affecting the conditions under which diffusion-based reconstruction operates. The decoder must therefore incorporate compression- and rescaling-aware conditioning to ensure consistency with the encoded representation and prevent uncontrolled generative artifacts.
Third, reconstruction difficulty varies across compression–rescaling combinations. Aggressive settings demand stronger generative enhancement for realistic detail recovery, whereas moderate settings require faithful content preservation. A unified model must therefore dynamically regulate the balance between fidelity and generative enhancement under diverse operating conditions.

% \IEEEpubidadjcol

In this paper, we address these challenges by proposing a diffusion-based joint arbitrary-scale super-resolution enhanced compression framework, termed \textbf{ASSR-EIC}. The central idea is to formulate variable ultra-low bitrate reconstruction as a degradation-aware generative modeling problem, conditioning diffusion on encoder-side compression and rescaling signals to ensure alignment with the transmitted representation. This design enables controllable detail restoration under varying degradation levels within a unified framework. 

Fig. \ref{fig:teaser} illustrates the overall architecture of ASSR-EIC. Built upon a fundamental anchor codec, either traditional or learning-based, that operates reliably at normal bitrates, the proposed framework enhances ultra-low bitrate performance while maintaining compatibility with existing systems. On the encoder side, an arbitrary-scale downsampling module enables flexible bitrate reduction. On the decoder side, diffusion-based ASSR reconstruction restores high-quality images at the original resolution.

To mitigate the ill-posed reconstruction problem caused by severe information loss, ASSR-EIC leverages diffusion-based generative priors to compensate for missing structural and textural constraints. To resolve the representation-alignment challenge introduced by encoder-side downsampling, ASSR-EIC incorporates compression- and rescaling-aware conditioning mechanisms that explicitly guide the generative process to remain consistent with the encoded representation. We observe that the compression quality parameter serves as an indicator of compression-induced distortion severity, while the rescaling factor reflects the extent of spatial information attenuation introduced during encoding. Together, these factors provide structured cues about the underlying degradation characteristics. Conditioning the diffusion model on such degradation-aware signals enables adaptive generative behavior under severe compression while preserving faithful details when distortion is mild.

To accommodate diverse compression–rescaling combinations, ASSR-EIC further integrates global and local adaptation within the diffusion process. Global conditioning injects compression and rescaling cues into the multi-resolution features of the backbone U-Net and the fidelity module, regulating the overall diffusion dynamics. Local modulation further adjusts the trade-off between generative enhancement and fidelity preservation. Together, these designs enable bitrate-adaptive reconstruction within a unified generative framework.

The main contributions can be summarized as follows:

\begin{itemize}
\item We extend joint SR-enhanced compression by integrating diffusion-based arbitrary-scale super-resolution (ASSR), enabling flexible and controllable variable-bitrate adaptation in the ultra-low bitrate regime within a unified and codec-compatible framework.

\item We introduce a degradation-aware diffusion reconstruction paradigm that models compression strength and rescaling factors as structured degradation descriptors, enabling explicit alignment between encoder-induced information loss and generative diffusion priors for adaptive fidelity–enhancement regulation.

\item Extensive experiments demonstrate that ASSR-EIC achieves state-of-the-art performance under variable ultra-low bitrate settings with controllable rate adaptation.
\end{itemize}

\section{Related Work}
\subsection{Lossy Image Compression} 
Lossy image compression plays a critical role in efficient image storage and transmission. Traditional codec standards, such as JPEG \cite{wallace1992jpeg}, HEVC \cite{sullivan2012overview}, and VVC \cite{JVET_VTM}, have been widely adopted in practice. However, these codecs often suffer from noticeable block artifacts, primarily due to limited spatial correlation modeling across adjacent image blocks.
% BPG \cite{BPG}

With the advent of deep learning, learned image compression has made remarkable progress. Ballé et al. \cite{balle2017end} first introduced neural networks into transform coding, followed by extensive studies \cite{ballé2018variational, minnenjoint, cheng2020learned, tian2020just, he2022elic, li2023neural, lu2025learned, zhu2022transformerbased, qian2022entroformer, koyuncu2022contextformer, liu2023learned} to achieve more accurate distribution estimation and improved compression efficiency. Nevertheless, as these approaches typically optimize the rate–distortion function with distortion-oriented losses such as PSNR or MS-SSIM, they tend to converge to mean predictions, leading to overly smooth reconstructions and perceptual blurriness.
% Cheng et al. \cite{cheng2020learned} propose a more accurate and flexible entropy model by employing discretized Gaussian mixture likelihoods. He et al. \cite{he2022elic} propose uneven channel-conditional adaptive coding to enhance compression performance with high efficiency. Liu et al. \cite{liu2023learned} develop a framework with parallel Transformer–CNN mixture (TCM) blocks that effectively combine the local modeling capacity of CNNs with the non-local modeling capacity of transformers. Lu et al. \cite{lu2025learned} introduce a dictionary-based cross-attention entropy model that leverages a learnable dictionary to summarize typical structures in the training data, thereby improving the entropy model.
Some recent methods exploit prior knowledge to improve perceptual quality. Pan et al. \cite{pan2024jnd} leverages the just noticeable difference (JND) as perceptual prior knowledge to effectively eliminate the perceptual redundancies in an image. Hu et al. \cite{hu2025text} incorporate textual descriptions into ROI-based image compression to achieve region-adaptive coding.

% It typically utilizes generative models to synthesize realistic details, with representative approaches based on GANs \cite{Agustsson_2019_ICCV,mentzer2020high,raman2020compressnet,iwai2021fidelity,agustsson2023multi,muckley2023improving}, VQ-VAE \cite{muckley2023improving,jia2024generative,li2025onceforall} and diffusion models \cite{theis2022lossy,ghouse2023residual,yang2023lossy,lei2023text+,li2024MISC,careil2024towards,li2024towards,kuang2024consistency,xiadiffpc}. 
To improve reconstruction realism, generative models have been incorporated into compression frameworks, giving rise to Generative Image Compression (GIC). 
Early studies primarily employed Generative Adversarial Networks (GANs) \cite{Agustsson_2019_ICCV, mentzer2020high, raman2020compressnet, iwai2021fidelity}. Mentzer et al. \cite{mentzer2020high} introduce HiFiC, which combines a conditional GAN with a hyperprior-based codec, achieving superior perceptual quality alongside substantial rate savings. Agustsson et al. \cite{agustsson2023multi} propose a beta conditioning mechanism to unify generative and non-generative representations, thereby enabling receiver-side control over the realism–distortion trade-off. MS-ILLM \cite{muckley2023improving} enhances statistical fidelity by employing a non-binary adversarial discriminator conditioned on VQ-VAE–quantized local representations.
% The development of diffusion models \cite{sohl2015deep,ho2020ddpm,songdenoising,dhariwal2021diffusion,nichol2022glide,rombach2022high} has substantially advanced image synthesis and spurred research on diffusion-based GIC. 
More recently, diffusion models have been increasingly applied to GIC. Theis et al. \cite{theis2022lossy} introduce an unconditional diffusion-based lossy compression framework, where Gaussian-noised pixel communication is employed to constrain the information rate. Yang et al. \cite{yang2023lossy} propose CDC, an end-to-end diffusion-based lossy image compression framework that conditions the diffusion reconstruction process with the encoded contextual latent.

\subsection{Ultra-low Bitrate Image Compression} 
Ultra-low bitrate image compression has attracted growing attention in recent years, targeting extremely low bitrates, often below 0.1 bpp. Gao et al. \cite{gao2023extremely} exploit the information-preserving property of invertible neural networks (INNs) to mitigate information loss in EIC. Jiang et al. \cite{jiang2023multi} incorporate text descriptions as semantic guidance within a GAN framework to enhance coding performance. Jia et al. \cite{jia2024generative} perform transform coding in the generative latent space of a VQ-VAE, enabling reconstructions that better align with human perception.

% The development of diffusion models \cite{sohl2015deep,ho2020ddpm,songdenoising,dhariwal2021diffusion,nichol2022glide,rombach2022high} 
Recently, diffusion-based methods have shown remarkable progress in ultra-low bitrate compression.  
%\cite{lei2023text+,li2024MISC,careil2024towards,li2024towards,kuang2024consistency,xiadiffpc}.
% Benefiting from the powerful generative diffusion prior \cite{rombach2022high}trained on billion-scale datasets, recent diffusion-based approaches have demonstrated substantial advances in ultra-low bitrate image compression. 
Text+Sketch \cite{lei2023text+} is an early approach that compresses text prompts and binary contour sketches at the encoder and uses them as inputs to a pre-trained text-to-image diffusion model for reconstruction at the decoder. MISC \cite{li2024MISC} leverages a Large Multimodal Model (LMM) to provide detailed text descriptions and positional maps as guidance for better reconstruction fidelity. PerCo \cite{careil2024towards} conditions the diffusion process on a vector-quantized latent image representation along with a textual image description. DiffEIC \cite{li2024towards} combines a VAE-based latent feature-guided compression module with diffusion priors to reconstruct images.
% Kuang et al. \cite{kuang2024consistency} incorporate an additional encoder following \cite{gao2023implicit} in the diffusion model to apply consistency guidance.
DiffPC \cite{xiadiffpc} proposes a two-stage framework and introduces a hybrid semantic refinement module to improve perceptual fidelity. 
% RDEIC \cite{li2025rdeic} proposes a relay residual diffusion that combines the residual diffusion with the pretrained diffusion prior for better fidelity and efficiency. 
Xu et al. \cite{xu2025decouple} masks latent representations based on the prior ROI information and
allocates limited bit resources to key areas to achieve
endogenous-adaptive perceptual image compression. PICD \cite{xu2025picd} proposes a versatile perceptual image compression framework for both screen and natural images by leveraging text content priors together with domain-, adaptor-, and instance-level conditioning.
% However, these methods lack rate control and require retraining multiple models for different R-D points to vary rates, resulting in significant training costs.

\subsection{Joint SR-Enhanced Compression and SR}
Instead of compressing the original image at high-resolution, joint SR-enhanced image compression attempt to use some pre-processing strategies to generate a low-resolution image for bit-rate savings. 
% Image rescaling is a joint process that downsamples an HR image into an LR version and then reconstructs the HR image from the LR version. 
Bruckstein et al. \cite{bruckstein2003down} pioneer the idea of incorporating the downsampling operation before compression to reduce bitrates, and reconstruct HR image at the decoder end. Subsequent studies improve the handcrafted downsampling \cite{lin2006adaptive} and upsampling \cite{wu2009low} strategies to enhance the performance. With the advent of deep learning, several approaches \cite{li18convolutional,afonso19video,tian2021self,amirpour2022deepstream,son2021enhanced,luo2022livesr,jeong2024real,zhao2025instance} combine CNN-based SR to reconstruct high-resolution images during decoding. 
% In \cite{son2021enhanced}, Son et al. propose an auxiliary codec to mimic JPEG, which is nondifferentiable, thereby providing more accurate gradients for optimization.
In \cite{jeong2024real}, Jeong et al. propose a real-time CNN training and compression method that delivers a low-resolution video segment and the corresponding compressed CNN parameters under a live streaming environment.
In \cite{zhao2025instance}, Zhao et al. leverage spatial-temporal enhancement with instance adaptation for efficient video compression.
AIDN \cite{xing2023scale} introduces a Conditional Resampling Module in a CNN encoder–decoder to enable invertible image downscaling with arbitrary scale factors by conditioning the resampling kernels on both scale and image content.

However, these methods operate only within regular bitrate ranges and struggle to handle ultra-low bitrate scenarios with severe information loss, primarily due to their CNN-based architectures lacking the generative ability. Recently, diffusion-based SR methods \cite{wang2024exploiting,lin2023diffbir,niu2024acdmsr,wu2024seesr,qu2024xpsr,chen2025faithdiff,sun2025pixel} have shown impressive progress. SeeSR \cite{wu2024seesr} trains a degradation-aware prompt extractor to generate soft and hard semantic prompts. XPSR \cite{qu2024xpsr} leverages MLLM to capture both high-level and low-level semantic prompts. FaithDiff \cite{chen2025faithdiff} develops an alignment module for faithful SR. PISA \cite{sun2025pixel} proposes a dual-LoRA diffusion method for pixel- and semantic-level adjustable SR. TADM \cite{wang2025timestep} proposes a diffusion-based latent rescaling framework for extreme image rescaling, introducing timestep-aware alignment to better adapt generative capacity under large scale factors.
% Wei et al. \cite{wei2024toward} propose a VQ-GAN–based rescaling method that reconstructs images with 16× or 32× downsampling. 
Inspired by these recent advances, we leverage the diffusion prior and introduce arbitrary-scale rescaling into the compression framework to achieve variable-rate ultra-low bitrate image compression.
% 下采样和上采样联合优化: invertible那篇
% 存在的问题，不在ulrta-low bitrate
% 缺乏scale可控
% extreme rescaling also shows potential to save bitrate（直接简单引用一下好了，不展开讲也不比了）；缺乏和image compression的结合，仅传输了downsampling的
% post-processing的research attention更多
% 从CNN-based BasicVSR到Transformer-based?看看其他文章怎么写的
% 最近基于Diffusion的realistic SR取得了长足的进展；和compression/rescaling的关系(微妙，看看别的文章怎么把握)

% Recent SR methods \cite{wang2024exploiting, lin2023diffbir, yang2025pixel,wu2024seesr,dong2025tsd} have shown impressive progress. They exploit the Text-to-Image (T2I) diffusion prior and design various attentions to improve the reconstruction quality. 
% DA-CLIP \cite{luo2023controlling} introduces a CLIP-based controller to automatically identify degradation types and perform corresponding image restoration.

% The success of recent diffusion-based image restoration methods inspires us to further compress image resolution during encoding and reconstruct it during decoding.
% and reconstructing high-resolution images at the decoding end.
% In this paper, we transfer the success of the realistic diffusion-based SR to joint compressed image restoration and continuous-scale image SR for extending existing codecs to ultra-low bitrates with high quality and high realism.

\begin{figure*}[h!]
  \centering
   \includegraphics[width=\linewidth]{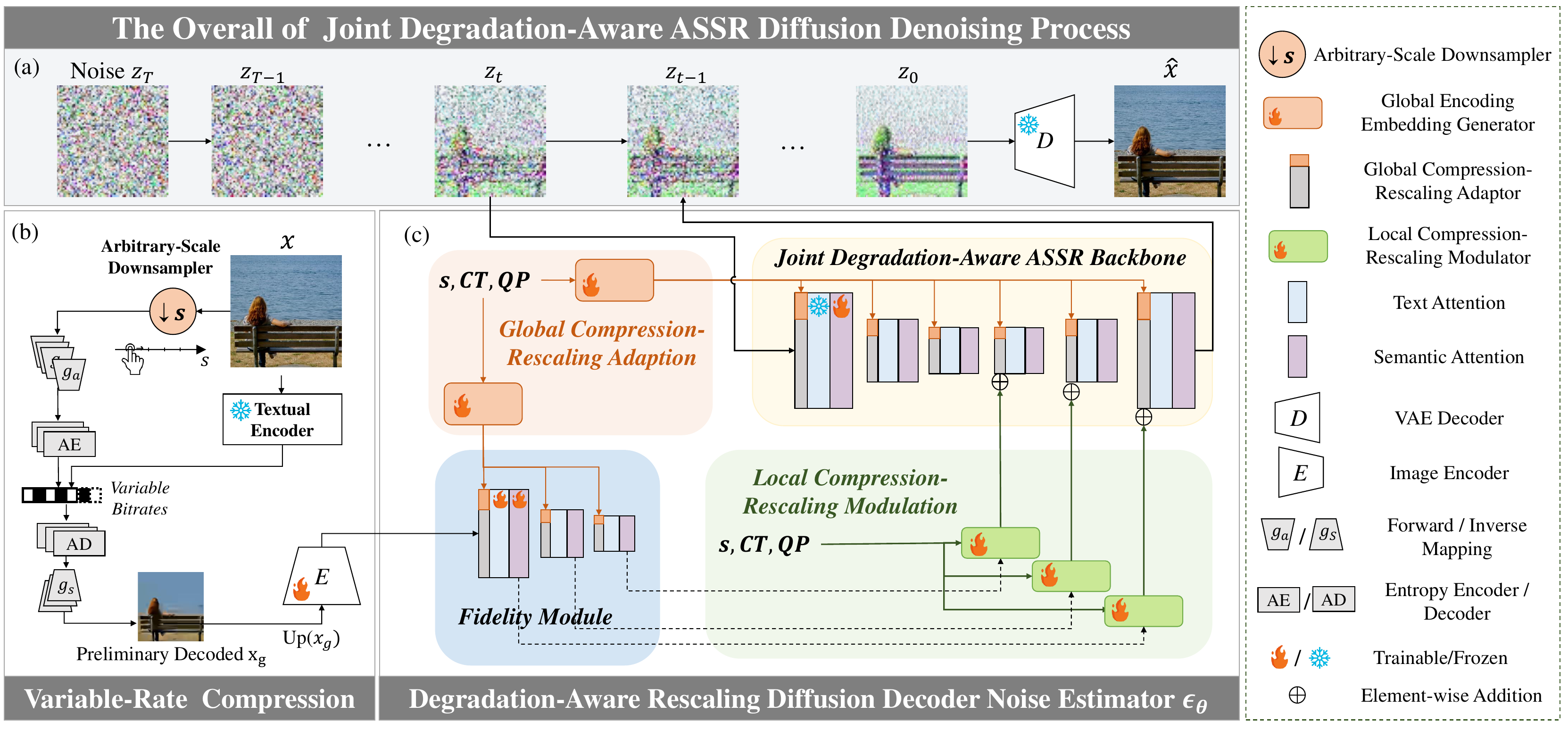}

   \caption{Structure overview of the proposed ASSR-EIC. We insert an arbitrary-scale downsampler to achieve controllable bitrate reduction at the encoder end, and propose a joint degradation-aware ASSR reconstruction decoder to restore the high-quality HR image with rate adaptation. The joint degradation-aware ASSR reconstruction decoder consists of a backbone, a fidelity module, a global compression-rescaling adaptor, and a local compression-rescaling modulator. The layers of the Backbone and the Fidelity Module are arranged from left to right to indicate the forward feature propagation order, with arrows omitted for clarity. $s, \chi_{CT}, \chi_{QP}$ denote the rescaling factor, codec type, and codec quality parameter, respectively.}
   \label{fig:over}
   % \vspace{-1em}
\end{figure*}

\subsection{Variable-Rate Image Compression} 
Variable-rate image compression aims to enable flexible bitrate adaptation within a single model to meet different practical bandwidth-limited requirements.
Existing variable-rate approaches can be broadly categorized into several groups. Progressive compression learns a scalable bitstream that can be incrementally decoded to support bitrates ranging from low to high \cite{toderici2017full,Johnston_2018_CVPR,Jeon_2023_CVPR,zhang2024exploring}. A second category achieves variable-rate compression by parameter conditioning, where  scalar parameters or conditional convolutions are introduced to dynamically adjust the quantization or coding process \cite{toderici2017full,Choi_2019_ICCV,li2025lightweight}. LERAN \cite{li2025lightweight} focuses on rate-adaptive compression-aware image rescaling by introducing quality factor-driven feature modulation, enabling a single lightweight model to handle multiple JPEG compression rates. Another line of research leverages multi-scale representations or slimmable networks \cite{cai2019efficient,Yang_2021_CVPR}. Other emerging approaches include context-based trit-plane coding \cite{Lee_2022_CVPR}, receptive field with uncertainty guidance \cite{zhang2024exploring} with neural codecs.

The reliance on diffusion models in recent ultra-low-bitrate methods substantially increases the training and deployment costs when multiple target-bitrate models are required. Nevertheless, achieving variable-bitrate EIC remains a challenge. To address this issue, we propose a novel variable-rate EIC framework that enables controllable rate adaptation through joint ASSR reconstruction.

% To tackle this challenge, some variable-rate neural image compression methods have been explored \cite{
% song2021variable,mei2021learning,bai2021learning,yang2021slimmable,lee2022dpict,choi2022scalable,zhang2023exploring,jeon2023context,guo2023toward,iwai2024controlling,liu2024rate,li2025onceforall}. Iwai et al. \cite{iwai2024controlling} explore several discriminator designs tailored for the variable-rate approach and introduce a novel adversarial loss. 
% Li et al. \cite{li2025onceforall} correlate the information density of local image patches with their granular representations and propose a VQGAN-based method. However, these methods are designed for normal bitrates and fail to achieve perceptual quality and image realism at ultra-low bitrates.
% In this paper, we address ultra-low bitrate scenarios and propose a compression framework that provides continuous rate control at ultra-low bitrates.

% In this paper, we unify ultra-low-bitrate compression and normal-bitrate compression in a shared framework by designing a plug-and-play versatile decoder.

\section{Method}

\subsection{Overview}
ASSR-EIC is designed for variable-bitrate EIC via incorporating joint compression-aware arbitrary-scale rescaling reconstruction. 
As illustrated in Fig. \ref{fig:over} (b), the proposed framework first spatially downsamples the high-resolution (HR) image to its low-resolution (LR) counterpart using an arbitrary-scale downsampling factor $s$, thereby controllably reducing data volume and bitrate. After the preliminary decoding of the LR image, we reconstruct the HR image using our proposed compression-aware diffusion-based arbitrary-scale super-resolution (ASSR) decoder. 

Our ASSR-enhanced image compression framework is compatible with both traditional codecs and learning-based image compression methods. Building upon an existing anchor codec, we insert an arbitrary-scale downsampler at the encoder end, while proposing a degradation-aware ASSR processing decoder at the decoder end.

Our method section is organized as follows. Section \ref{sec:rescaling} presents the proposed joint ASSR-enhanced variable-bitrate compression framework. Section \ref{sec:CRD} details the diffusion-based, degradation-aware ASSR decoder. Sections \ref{sec:adaptor} and \ref{sec:modulator} further describe the design of the variable-rate adaptive reconstruction modules.

% \begin{table*}[!t]
% \centering
% \renewcommand{\arraystretch}{1.5}
% \setlength{\tabcolsep}{10pt}
% \caption{Impact of Spatial Downsampling on Bitrate for Representative Anchor Codecs. 
% The value in parentheses denotes the ratio 
% $\mathrm{bpp}_{\mathrm{orig}} / \mathrm{bpp}_{\mathrm{down}}$, 
% indicating how many times the bitrate is reduced after spatial downsampling.}
% \label{tab:bpp_downsampling}
% \begin{tabular}{@{}lcccccc@{}}
% \toprule
% \textbf{Codec} & \textbf{Quality Parameter} & \textbf{bpp (Orig.)} & \textbf{bpp ($s{=}1.2$)} & \textbf{bpp ($s{=}1.5$)} & \textbf{bpp ($s{=}1.8$)} & \textbf{bpp ($s{=}2$)} \\
% \midrule

% VVC    & QP=47 & 0.124 & 0.080\,(1.55$\times$) & 0.059\,(2.10$\times$) & 0.047\,(2.64$\times$) & 0.040\,(3.10$\times$) \\
%        & QP=52 & 0.057 & 0.042\,(1.36$\times$) & 0.033\,(1.73$\times$) & 0.027\,(2.11$\times$) & 0.024\,(2.38$\times$) \\

% MS-ILLM & "quality\_2" & 0.091 & 0.069\,(1.32$\times$) & 0.054\,(1.69$\times$) & 0.041\,(2.22$\times$) & 0.029\,(3.14$\times$) \\
%        & "quality\_1"  & 0.053 & 0.042\,(1.26$\times$) & 0.033\,(1.61$\times$) & 0.025\,(2.12$\times$) & 0.019\,(2.79$\times$) \\

% \midrule
% \multicolumn{2}{c}{\textbf{Avg.}$^{\ast}$} & -- 
% & 1.37$\times$ ($\approx 0.95\, s^2$) 
% & 1.78$\times$ ($\approx 0.79\, s^2$) 
% & 2.27$\times$ ($\approx 0.70\, s^2$)
% & 2.85$\times$ ($\approx 0.71\, s^2$) \\
% \bottomrule
% \end{tabular}

% \vspace{2pt}
% {\footnotesize
% \noindent
% $^{\ast}$The bitrate reduction factor increases as the downsampling scale grows or the original bitrate decreases.\\
% }
% \end{table*}

\begin{table*}[!t]
\centering
\renewcommand{\arraystretch}{1.5}
\setlength{\tabcolsep}{15pt}
\caption{Impact of Spatial Downsampling on Bitrate for Representative Anchor Codecs on MSCOCO. 
The value in parentheses denotes the ratio 
$\mathrm{bpp}_{\mathrm{orig}} / \mathrm{bpp}_{\mathrm{down}}$, 
indicating how many times the bitrate is reduced after spatial downsampling.}
\label{tab:bpp_downsampling}
\begin{tabular}{@{\hspace{6pt}}lccccc@{\hspace{6pt}}}
\toprule
\textbf{Codec} & \textbf{Quality Parameter} & \textbf{bpp (Orig.)} & \textbf{bpp ($s{=}1.2$)} & \textbf{bpp ($s{=}1.5$)} & \textbf{bpp ($s{=}2$)} \\
\midrule

VVC    & QP=47 & 0.124 & 0.080\,(1.55$\times$) & 0.059\,(2.10$\times$) & 0.040\,(3.10$\times$) \\
       & QP=52 & 0.057 & 0.042\,(1.36$\times$) & 0.033\,(1.73$\times$) & 0.024\,(2.38$\times$) \\

MS-ILLM & "quality\_2" & 0.091 & 0.069\,(1.32$\times$) & 0.054\,(1.69$\times$) & 0.029\,(3.14$\times$) \\
       & "quality\_1"  & 0.053 & 0.042\,(1.26$\times$) & 0.033\,(1.61$\times$) & 0.019\,(2.79$\times$) \\

\midrule
\multicolumn{2}{c}{\textbf{Avg.}$^{\ast}$} & -- 
& $\approx$1.37$\times$ ($\approx 0.95\, s^2$) 
& $\approx$1.78$\times$ ($\approx 0.79\, s^2$) 
& $\approx$2.85$\times$ ($\approx 0.71\, s^2$) \\
\bottomrule
\label{tab:bpp}
\end{tabular}

\vspace{-3pt}
{\footnotesize
\parbox{\linewidth}{
$^{\ast}$The empirical bitrate reduction demonstrates an approximate quadratic trend with respect to the downsampling scale. Deviations from ideal $s^2$ scaling become more noticeable under larger scaling factors or more aggressive compression settings.
}
}
\vspace{-1em}
\end{table*}

\subsection{Joint ASSR-Enhanced Image Compression Framework} \label{sec:rescaling}

Following the idea of joint SR-enhanced compression \cite{khani2021efficient,tian2021self,li2022efficient,amirpour2022deepstream,luo2022livesr,jeong2024real,zhao2025instance,yu2025stsr360}, spatial downsampling can be introduced prior to encoding so that the codec operates on a lower-resolution representation, thereby reducing the effective bitrate. Building upon this paradigm, we construct a plug-and-play ASSR-enhanced compression framework on top of an existing anchor codec. For notational clarity, the encoding-decoding pipeline of the anchor codec is abstracted in Fig. \ref{fig:over} as a pair of forward and inverse mappings, denoted by $g_a$ and $g_s$, along with the entropy coding and decoding components (AE and AD). Together, these modules represent the original codec process, which remains unchanged in our design. Our framework operates externally by inserting an arbitrary-scale spatial downsampler before the encoding stage and introducing a degradation-aware ASSR diffusion decoder after decoding, without modifying any internal components of the anchor codec.

To empirically examine the impact of spatial downsampling on compression rate under ultra-low bitrate conditions, we analyze two representative anchor codecs, the traditional VVC and the learning-based MS-ILLM. As shown in Tab.~\ref{tab:bpp}, on average, the empirical reduction exhibits an approximate quadratic trend with respect to the downsampling scale. Although the reduction does not strictly follow ideal quadratic ($s^2$) scaling, particularly at larger downsampling factors or under more aggressive compression settings, substantial rate savings are consistently observed. The deviation from ideal scaling is influenced by two main factors. Larger downsampling modifies the signal statistics presented to the encoder and may increase the relative proportion of high-frequency components in the reduced-resolution representation, which can affect coding efficiency under the same compression configuration. In addition, at extremely low bitrates, the remaining redundancy becomes limited, leading to diminishing marginal rate gains.

% \textcolor{blue}{While spatial downsampling provides an effective mechanism for bitrate reduction, it inevitably increases reconstruction difficulty, particularly under aggressive compression settings. Structural and textural information becomes increasingly attenuated as compression and rescaling accumulate, making high-quality recovery substantially more challenging.}

% \textcolor{blue}{To address this challenge, we introduce a degradation-aware ASSR reconstruction decoder at the decoder side. By explicitly modeling the combined effects of compression and rescaling within a unified reconstruction process, the framework enables controllable bitrate reduction while maintaining structural fidelity and perceptual quality.}

Beyond handling aggressive bitrate reduction, the proposed design naturally supports a unified processing pipeline across different bitrate regimes. For ultra-low-bitrate compression, the full pipeline is activated, including spatial downsampling and the joint diffusion-based ASSR decoder. For normal-bitrate compression, the framework degenerates to the original anchor codec by bypassing the downsampling module and the diffusion decoder, or equivalently by setting the rescaling factor to 1, without modifying the codec structure itself:

\begin{equation}
\label{eq: framework}
    \begin{cases}
      s \geq 1,\ (CRD)^{on} & for \ Ultra\text{-}Low\ BitRate,\\
      s = 1,\ (CRD)^{off} & for \ Regular\ BitRate.
    \end{cases}
\end{equation}
where $CRD$ denotes the diffusion-based degradation-aware ASSR reconstruction decoder. This plug-and-play design enables seamless switching between operating conditions while preserving compatibility with existing codecs. The rescaling factor $s$ regulates bitrate at the signal level by modifying the spatial dimensionality prior to encoding, while the anchor codec continues to perform its native rate allocation on the rescaled representation. As a result, bitrate control is achieved through the joint effect of structural rescaling and codec-level quantization. Since $s$ can be adjusted at inference time, it expands the controllable bitrate range and, in combination with the codec’s native rate control, provides practical rate scalability across diverse operating conditions.

% For the ultra-low bitrate scenario, we set $s \geq 1$ and utilize the diffusion-based joint ASSR decoder to reconstruct high-quality images under extreme encoding conditions. While for regular bitrate ranges, we simply set the rescaling factor $s=1$ and disable the diffusion-based joint ASSR decoder, in which case the ASSR-EIC codec directly degenerates into the underlying anchor codec.

% For the choice of $s$, assume encoding an image $x$ requires a rate $R$. The rate $R_s$ for the downsampled image $x_s$ with the downsampling scale factor $s$ is typically given by:

% Considering the trade-off between information loss and bitrate saving, we set $s\leq2$ in this paper, which is sufficient to achieve ultra-low bitrates below 0.02 bpp.

% Assume the bitrate of the original image is $R_1$. When the image is downsampled by a scale factor $s$, the bitrate of the downsampled image becomes $R_2$. Typically, $R_2$ is slightly larger than $R_1$, but the actual bitrate required to compress the original image is reduced to $\frac{R_2}{s^2}$, yielding nearly a quadratic reduction in bitrate.

\subsection{Joint Degradation-Aware ASSR Decoder} \label{sec:CRD}
The key component of our framework is the diffusion-based joint degradation-aware ASSR decoder. It reconstructs high-quality, high-resolution decoded images under various compression settings and rescaling scales within a single unified diffusion model. 
% To achieve rate adaptation under different conditions, we incorporate both compression awareness and rescaling awareness into the reconstruction process.
% process on the low-resolution coarsely decoded reference image $x_g$, first-stage encoding parameters $QP$ and the rescaling factor $s$. 

As shown in Fig. \ref{fig:over} (b), given the preliminary decoded image $x_g$, similar to recent diffusion-based SR methods \cite{wu2024seesr,qu2024xpsr,chen2025faithdiff,sun2025pixel}, we first upsample $x_g$ to the original resolution and adopt a diffusion model to remove the compression artifacts and reconstruct the structural and textural details. To handle various compression and rescaling settings, we introduce compression awareness and rescaling awareness into the diffusion process, enabling effective rate adaptation. We adopt Stable Diffusion \cite{rombach2022high} as the main backbone of our framework and initialize it with pre-trained weights to leverage its diffusion prior for enhanced realism. We inherit the text attention module from Stable Diffusion and extract the original image's caption via a textual encoder as the prompt for the text attention module. 

To improve reconstruction fidelity, we further design a fidelity module that extracts structural and textural information from $x_g$. This module follows a ControlNet-like architecture \cite{zhang2023adding}. It extracts multi-resolution features from the preliminary decoded image $x_g$ and adds the features from the last layer at each resolution to the first-layer features of the corresponding resolution in the main backbone, as shown in Fig. \ref{fig:over} (c).

We also incorporate a dual semantic-enhanced design into both the diffusion backbone and the fidelity module, which provides additional semantic guidance and further improves reconstruction fidelity.
Firstly, the image caption prompt provides an overall semantic guidance by leveraging the pre-trained diffusion prior. Specifically, we extract an image caption from the original image $x$ during encoding and losslessly compress it using Lempel–Ziv coding. During decoding, the decompressed caption is employed as the diffusion prompt, providing global semantic guidance that remains clean and entirely unaffected by compression artifacts. Secondly, inspired by SeeSR \cite{wu2024seesr}, we extract image features as soft semantic prompts. SAM \cite{kirillov2023segment}, a general-purpose segmentation model, is inherently robust to compression distortions; therefore, we use the image encoder of SAM-Tiny to derive semantic features, which are integrated into the diffusion process through a dedicated semantic attention mechanism:

\begin{equation}
SeAttn(Q,K,V)=softmax(\frac{QK^T}{\sqrt{d_K}}) \cdot V
\end{equation}
where $d_K$ represents the dimension of the $K$ vector, and query $Q$, key $K$, value $V$ are obtained by:
\begin{equation}
    Q=W_Q^{(i)} \cdot F_{U}^{(i)}, K=W_K^{(i)} \cdot F_{sem}, V=W_V^{(i)} \cdot F_{sem}
\end{equation}
where $F_{U}^{(i)}$ denotes the intermediate feature from the $i$-th layer of the denoising UNet, which serves as the UNet-side input to the semantic attention module, $F_{sem}$ denotes the extracted SAM features, and $W_Q^{(i)}, W_K^{(i)}, W_V^{(i)}$ are learnable projection matrices.

\subsection{Global Compression-Rescaling Adaptor} \label{sec:adaptor}
To adaptively achieve rate adaptation, we explicitly introduce compression awareness and rescaling awareness into the diffusion process of the ASSR diffusion decoder. Specifically, we first achieve a global compression-rescaling awareness through a global adaptor, injecting both compression and rescaling information into the multi-resolution features of the backbone UNet and the fidelity module, thereby regulating the overall diffusion process.

Compression quality and rescaling factors during encoding characterize the severity of encoder-induced degradation. By determining the amount of information preserved in the transmitted representation, they directly affect reconstruction difficulty and thus provide structured conditioning signals for rate-adaptive generative modeling. We model three encoding variables: the codec type $\chi_{CT}$, codec quality parameter $\chi_{QP}$, and rescaling factor $s$. We set the codec type $\chi_{CT}$ to 0 for traditional codecs and 1 for learning-based codecs, motivated by the observation that traditional codecs typically exhibit structured artifacts such as blocking, whereas learning-based codecs commonly produce smoother or blur-like distortions, especially under pixel-wise optimization. This coarse categorization provides a simple yet effective inductive bias that helps reduce the learning complexity of the model. To align compression quality parameters across different codecs, we use the bitrate measured on the Kodak \cite{Franzen1999} dataset as a reference quality indicator $\chi_{QP}$, which serves as a unified proxy for codec quality comparison. The rescaling factor $s$ controls the spatial downsampling applied at the beginning of the pipeline.

\begin{figure}[t]
  \centering
   \includegraphics[width=\linewidth]{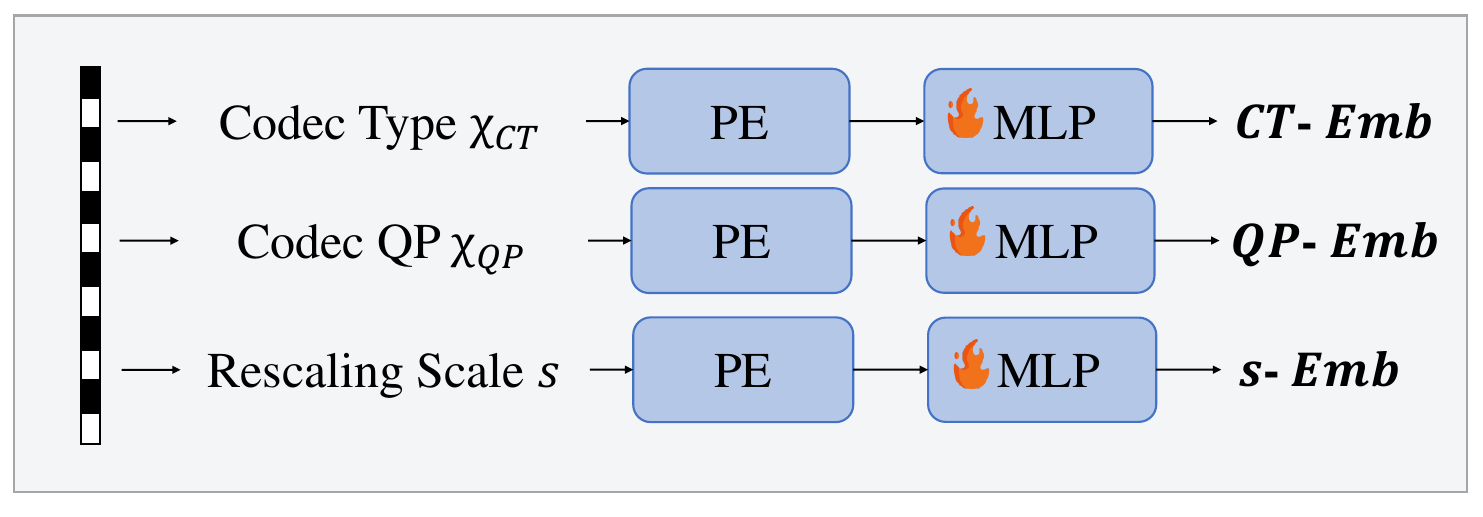}

   \caption{Details of the encoding embedding generation. We encode the float-type compression parameters and the rescaling factor into feature embeddings.}
   \label{fig:embed}
   % \vspace{-1em}
\end{figure}

\begin{figure}[t]
  \centering
   \includegraphics[width=\linewidth]{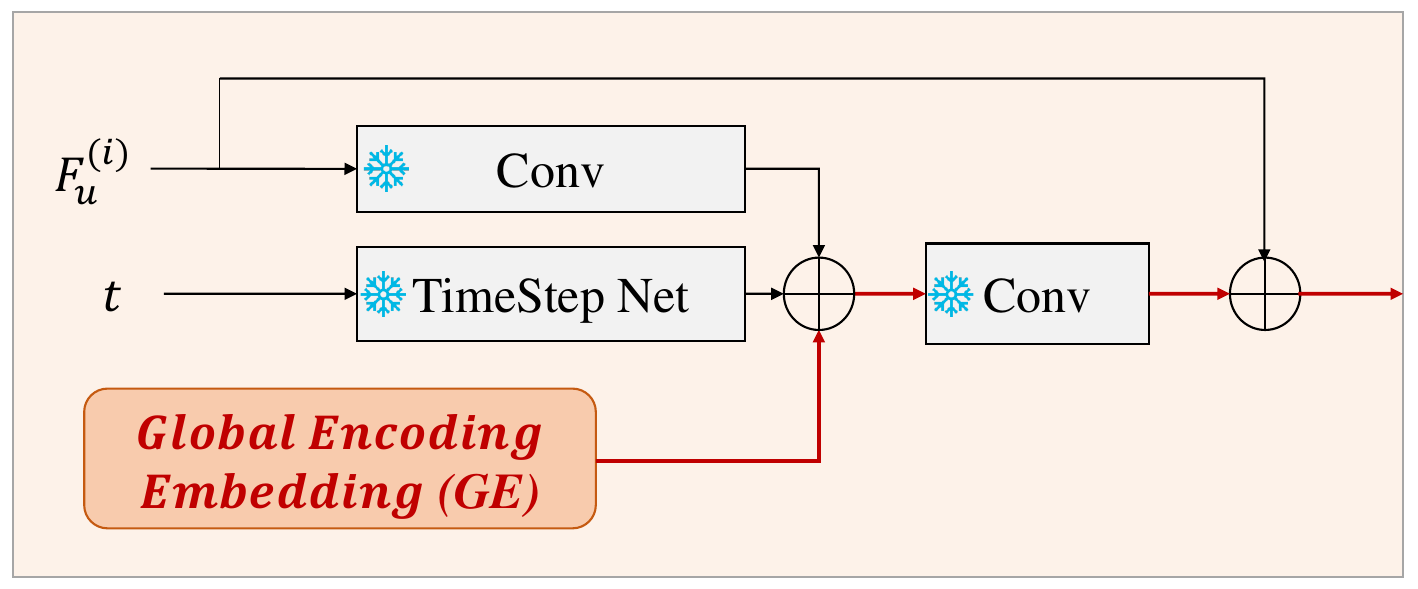}

   \caption{Detailed structure of the global compression-rescaling adaptor.}
   \label{fig:global}
   % \vspace{-1em}
\end{figure}

Inspired by the timestep incorporation approach of Stable Diffusion \cite{rombach2022high}, we first convert the float encoding parameters into encoding embedding. As shown in Fig. \ref{fig:embed}, we use position encoding and a trainable multi-layer perceptron (MLP) to obtain the embedding for each encoding parameter. Specifically, the embedding process is defined as follows:

\begin{equation}
    Emb{(p)} = Linear\left( SiLU\left( Linear(PE(p)) \right) \right), \\
\label{equa:embed}
\end{equation}
where 
$PE(p)$ represents the position encoding for the parameter $p$, $p=\chi_{CT},\chi_{QP},s$, and the subsequent operations involve two linear layers with a SiLU activation function.

% To better accommodate variable encoding conditions, we incorporate rate-specific control mechanisms to enable compression-rescaling-adaptive feature adjustment in both the backbone and the fidelity module, resulting in improved high-quality reconstruction. 
% Our rescaling diffusion decoder is designed to handle the complex and diverse mixture of distortions within a single model, including rescaling artifacts and compression artifacts from various compression methods with different quality factors. Since rescaling distortion is directly related to rate, it is crucial to decouple it from other distortions to enable more effective adaptation. To address this, we introduce the Rate-Distortion-Scale (RDS) Aligner. This module dynamically aligns the various distortion domains to the high-quality image domain based on the encoding parameters, enabling adaptive enhancement while isolating rescaling distortion from other compression-related artifacts.

To enable global adaptation of the diffusion process based on the encoding conditions, we introduce a global compression–rescaling adaptor. This module is built upon the original ResNet block of the Stable Diffusion backbone. As illustrated in Fig. \ref{fig:global}, the original ResNet block is kept frozen, preserving the pre-trained diffusion prior, while we inject an additional global encoding embedding to modulate the feature activations for global compression–rescaling awareness. The global encoding embedding is computed as:

\begin{align}
GE &= CT\text{-}Emb + QP\text{-}Emb + s\text{-}Emb, \\
F_G^{(i)} &= \mathrm{conv_2}\big(\mathrm{conv_1}(F_u^{(i)}) + TNet(t) + GE\big) + F_u^{(i)}.
\end{align}
where $GE$ denotes the fused global encoding embedding, which is added to the ResNet backbone before the fusion convolution, as shown in Fig.~\ref{fig:global}. 
$F_u^{(i)}$ denotes the input feature of the ResNet block in the global compression-rescaling adaptor at the $i$-th level of the denoising UNet, and $TNet(\cdot)$ denotes the timestep embedding network. $F_G^{(i)}$ denotes the corresponding output feature of the global compression-rescaling adaptor at level $i$.

We learn separate global encoding embeddings for the ASSR backbone and the fidelity module, allowing each to adapt to its respective functional role. In the ASSR backbone, all parameters of the underlying ResNet block are kept frozen, and only the global encoding embedding generation networks are trainable so as to preserve the diffusion prior. For the fidelity module, all parameters of the global compression–rescaling adaptor are trainable.

\subsection{Local Compression-Rescaling Modulator}  \label{sec:modulator}
While the global compression–rescaling modulator offers overall guidance for rate-adaptive reconstruction, accurately recovering fine-grained structures and textures is considerably more challenging and calls for precise local modulation. In particular, the emphasis of detail reconstruction varies across different encoding conditions, making high-quality detail recovery under rate adaptation even more difficult. To address this, we introduce a local compression–rescaling modulator between the fidelity module and the degradation-aware ASSR backbone, enabling adaptive control of detail synthesis under varying encoding conditions.

\begin{figure}[t]
  \centering
   \includegraphics[width=\linewidth]{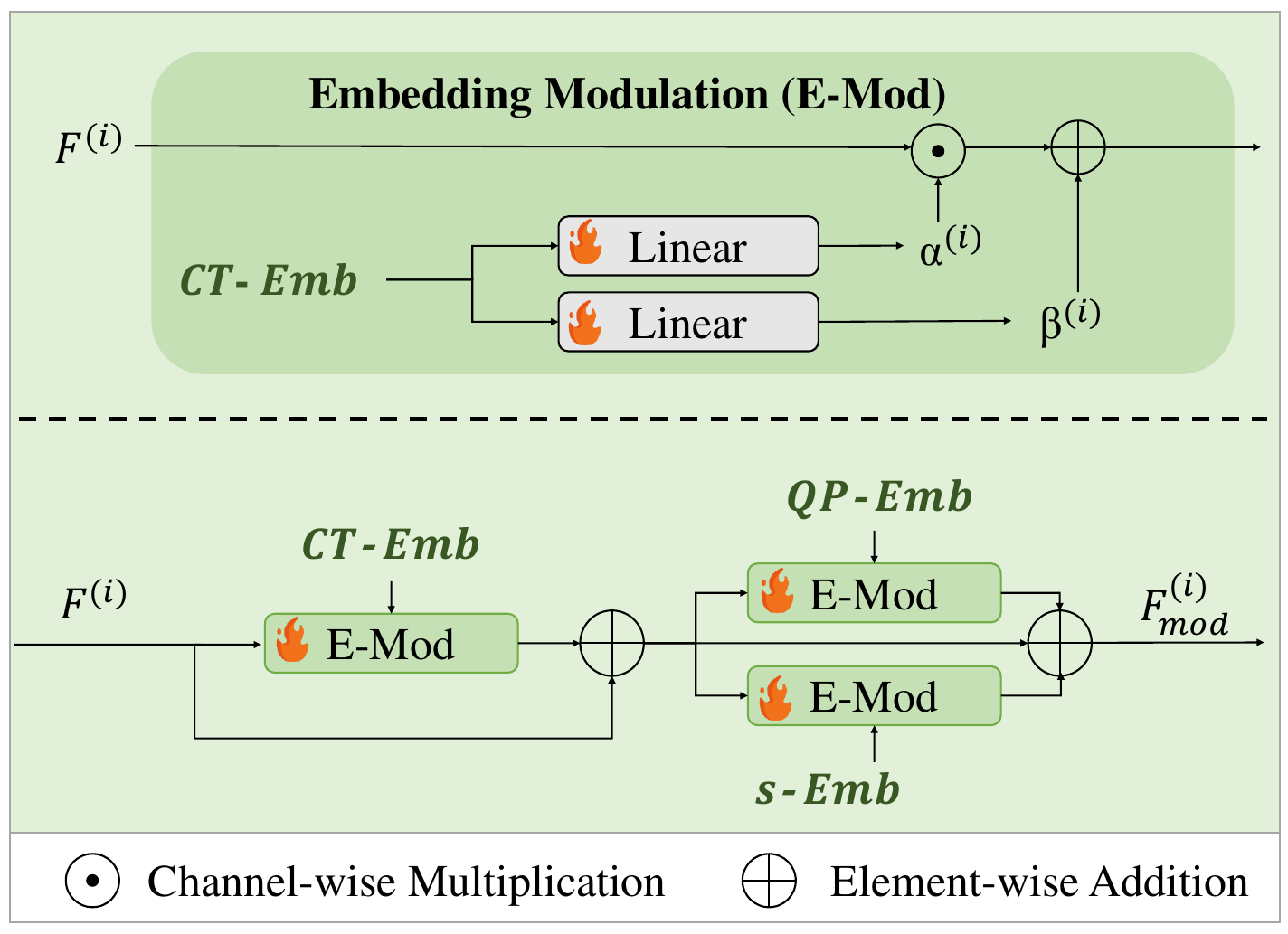}

   \caption{Detailed structure of the local compression-rescaling modulator.}
   \label{fig:Modulator}
   % \vspace{-1em}
\end{figure}

The local compression–rescaling modulator refines the multi-scale features from the fidelity module before they are fed into the degradation-aware ASSR backbone. 
Specifically, we first obtain the codec type embedding, codec quality parameter embedding, and rescaling scale embedding following the same embedding generation procedure described in Fig. \ref{fig:embed} and Equa. \ref{equa:embed}. These embeddings are then used to modulate the fidelity module features. As illustrated in Fig. \ref{fig:Modulator}, the embedding modulations are applied sequentially.Given the $i$-th level output feature $F^{(i)}$ from the fidelity module, the codec type embedding modulation is applied first, followed by two parallel embedding modulations using the codec quality parameter and the rescaling scale. The resulting output of the local compression–rescaling modulator is denoted as $F_{mod}^{(i)}$.

The embedding modulation (E-Mod) mechanism is shown at the top of Fig. \ref{fig:Modulator}. Given an embedding, such as $CT\text{-}Emb$, two linear layers generate a gain term $\alpha^{(i)}$ and a bias term $\beta^{(i)}$, which modulate the features in the $i$-th denoising UNet layer:
\begin{equation}
    \alpha^{(i)} = Linear^{(i)}_{\alpha}(CT\text{-}Emb),
\end{equation}
\begin{equation}
    \beta^{(i)} = Linear^{(i)}_{\beta}(CT\text{-}Emb),
\end{equation}
\begin{equation}
    F^{(i)}_{out} = \alpha^{(i)}\odot F^{(i)}+\beta^{(i)}
    \label{inter}
\end{equation}
where $\odot$ indicates channel-wise multiplication. $F^{(i)}_{out}$ represents the output feature after embedding modulation with $CT\text{-}Emb$. Similar embedding modulations are applied using $QP\text{-}Emb$ and $s\text{-}Emb$.

This compression-rescaling modulation mechanism refines the features from the fidelity module to adaptively reconstruct local details according to the specific compression and rescaling conditions, thereby improving the reconstruction performance across diverse operating conditions.

\subsection{Loss Function}
\textbf{Latent Diffusion Loss}:
To supervise the rescaling diffusion decoding process, we adopt a latent diffusion loss. During training, the original image $x$ is encoded into its latent representation $z_0$ via a pre-trained VAE encoder \cite{rombach2022high}. Noise is gradually added to $z_0$ to obtain a noisy latent $z_t$, where $t$ denotes the diffusion timestep. $\epsilon$ denotes the ground-truth Gaussian noise sampled from a standard normal distribution, which is added to the clean latent during the forward diffusion process. Given the timestep $t$, the noisy latent $z_t$, the image caption $c$, the semantic feature $F_{sem}$, and the encoding parameters $\chi_{CT}, \chi_{QP}, s$, our joint degradation-aware ASSR diffusion decoder learns to estimate the added noise via the noise prediction network $\epsilon_{\theta}$: 

\begin{equation}
\begin{split}
    L_{diff}=&\mathbb{E}_{z_0, z_t, t,c,F_{sem},\chi_{CT},\chi_{QP},s,\epsilon\sim N}[||\epsilon- \\
    &\epsilon_\theta(z_0,z_t,t,c,F_{sem},\chi_{CT},\chi_{QP},s)||_2^2]
\end{split}
\end{equation}

\textbf{Domain Alignment Loss}:
To align the latent domain encoded from the preliminary decoded image $x_g$,which suffers from quality degradation due to compression artifacts, with the high-quality latent domain of the pre-trained diffusion prior, we unfreeze the image encoder and supervise the intermediate features to encourage it to directly map to the high-quality latent domain. Specifically, similar to PASD \cite{sun2025pixel}, we apply convolution layers to convert the multi-scale intermediate image encoder features into corresponding-resolution
RGB images and compute the $L_1$ distance between the
converted RGB images and the downsampled corresponding-resolution ground-truth RGB images: 

\begin{equation} 
L_{A} = \sum_{n=1,2,3}||x^{(n)} - \text{toRGB}(f_n)||_1 
\end{equation} 
where $f_n$ represents the n-th level image encoder intermediate features, which are in 1/2, 1/4 and 1/8 scaled resolutions, and $x^{(n)}$ is the bicubic-downsampled ground-truth to match the spatial resolution of $f_n$. $toRGB$ refers to the convolution layers that convert the intermediate features into a 3-channel RGB image.

% \textbf{Perceptual Loss}:
% We adopt the perceptual loss $L_{LPIPS}$ \cite{zhang2018unreasonable} to encourage high-level perceptual quality and semantic consistency.

\textbf{Overall Loss}:
The total loss function is a weighted sum of the above losses: \begin{equation} 
L_{total} = L_A + L_{diff}
\end{equation} 

\begin{figure*}[htbp]
% \begin{minipage}[c]{\linewidth}
  \centering
   \includegraphics[width=\linewidth]{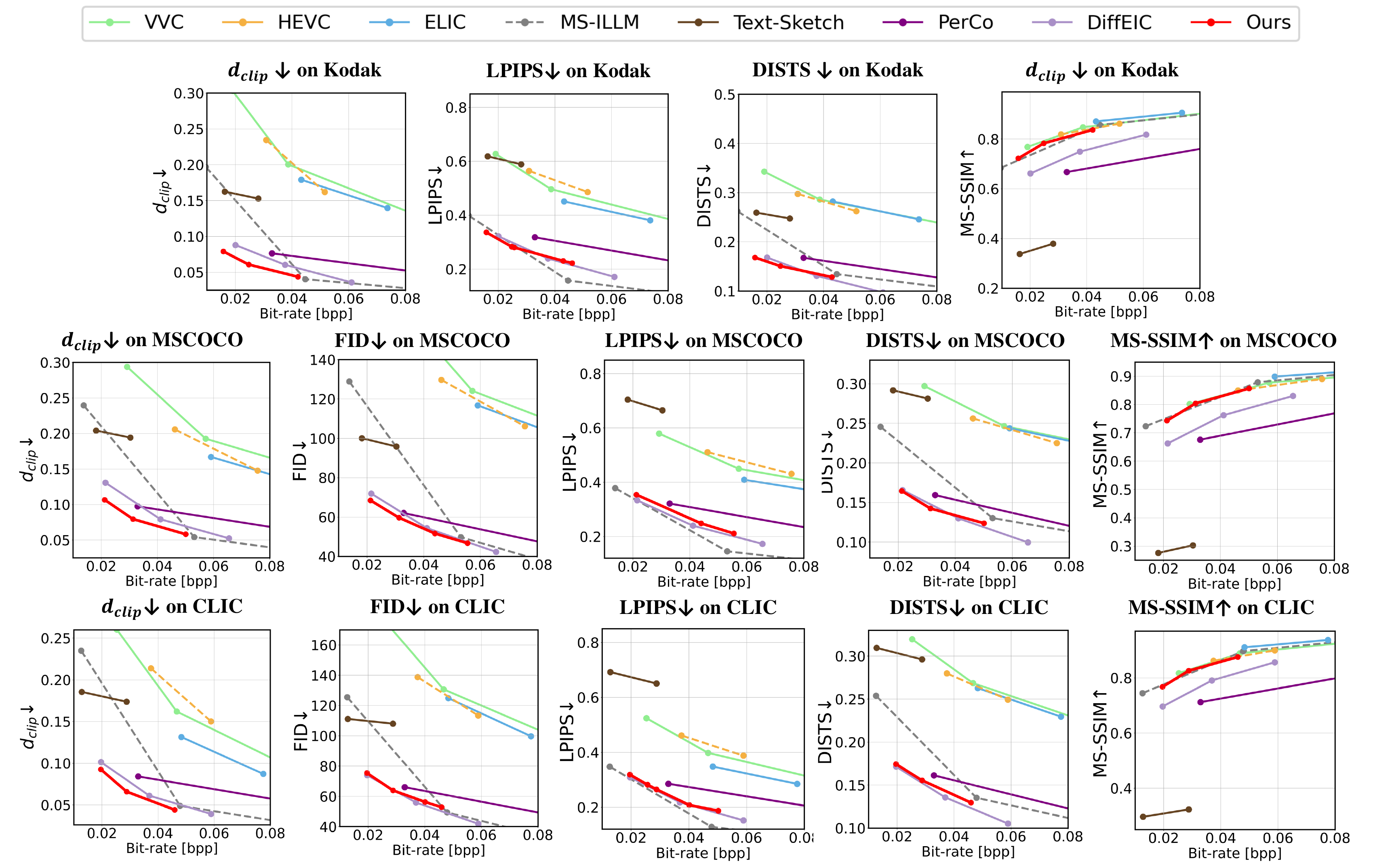}
   \caption{Evaluation of ASSR-EIC and competing compression codecs on the Kodak, MSCOCO, and CLIC2020 datasets. FID is not reported on Kodak because its limited dataset size is insufficient for reliable FID estimation.
   % 我们使用基于MS-ILLM的anchor的方法来展示我们方法的主要结果
   }
   \label{fig:codec_obj}
   \vspace{0.2em}
% \end{minipage}
\end{figure*}

\begin{figure*}[!t]
  \centering
   \includegraphics[width=\linewidth]{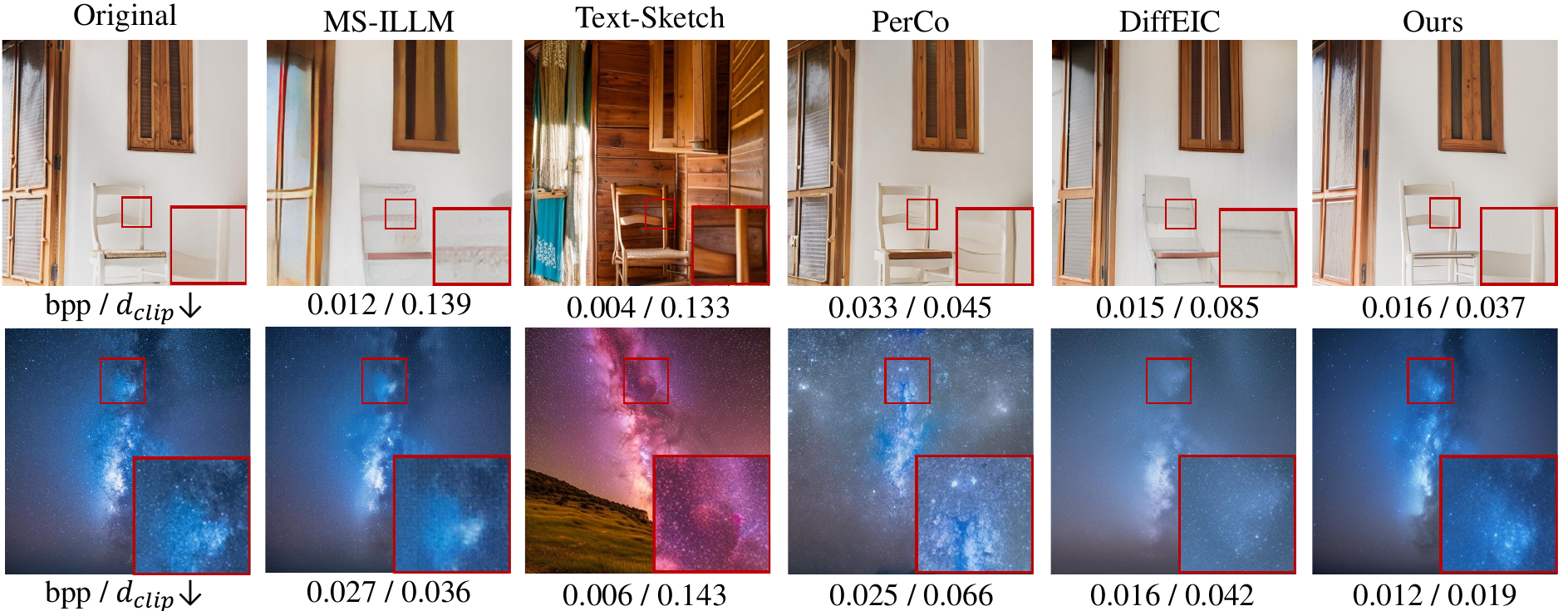}

   \caption{Visual comparison of our ASSR-EIC against representative state-of-the-art image compression codecs.
   % 我们使用基于MS-ILLM的anchor的方法来展示我们方法的主要结果
   }
   \label{fig:codec_subj}
%    \vspace{0.2em}
\end{figure*}

\begin{figure*}[!t]
% \begin{minipage}[c]{0.96\linewidth}
  \centering
   \includegraphics[width=\linewidth]{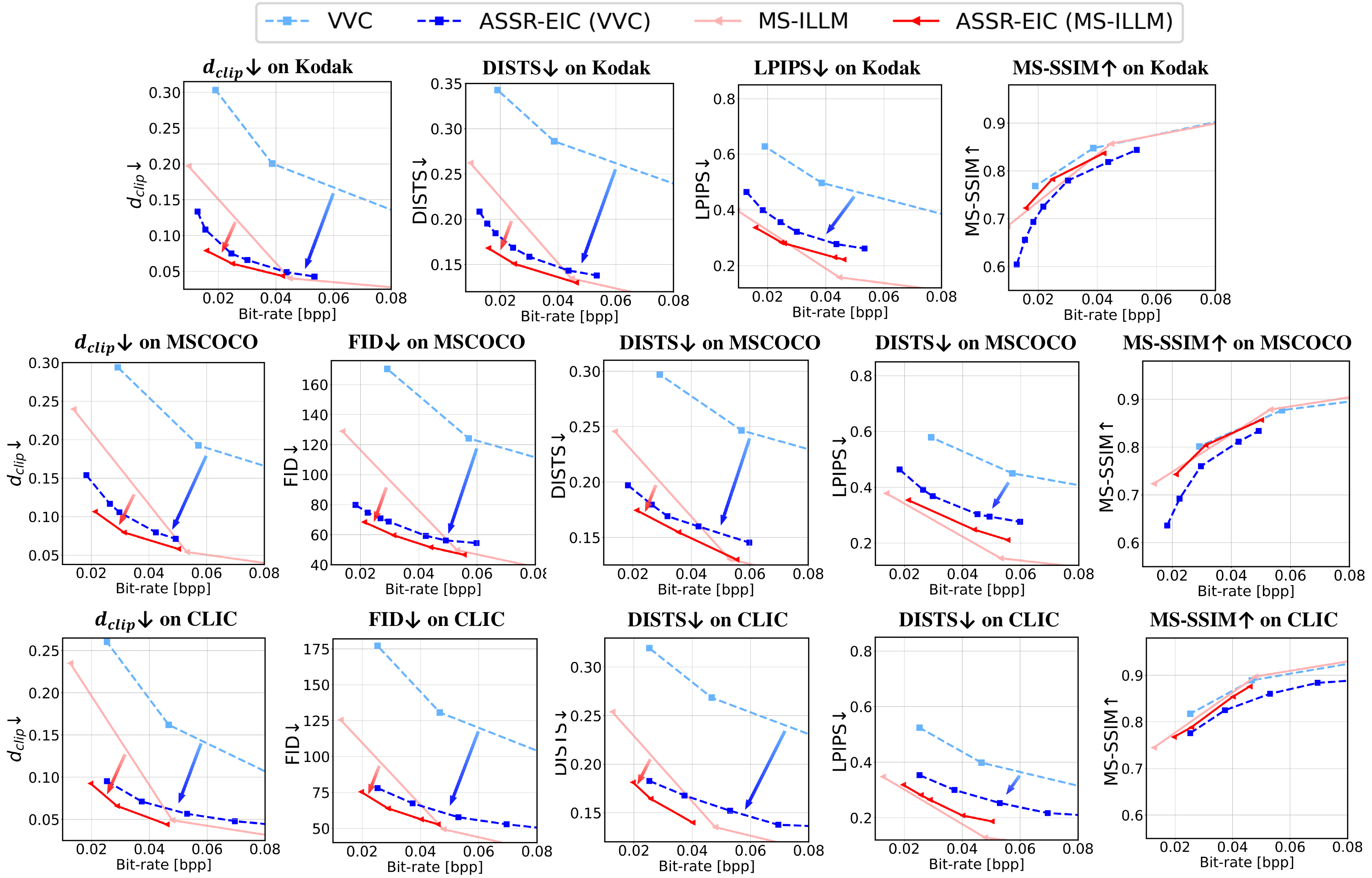}
   \caption{Rate–distortion performance improvements of our ASSR-EIC over the basic VVC and MS-ILLM anchors on the Kodak, MSCOCO, and CLIC2020. FID is not reported on Kodak because its limited dataset size is insufficient for reliable FID estimation.}
   \label{fig:codec_obj_gain}
   \vspace{-0.2em}
% \end{minipage}
\end{figure*}

\begin{figure*}[!t]
  \centering
   \includegraphics[width=\linewidth]{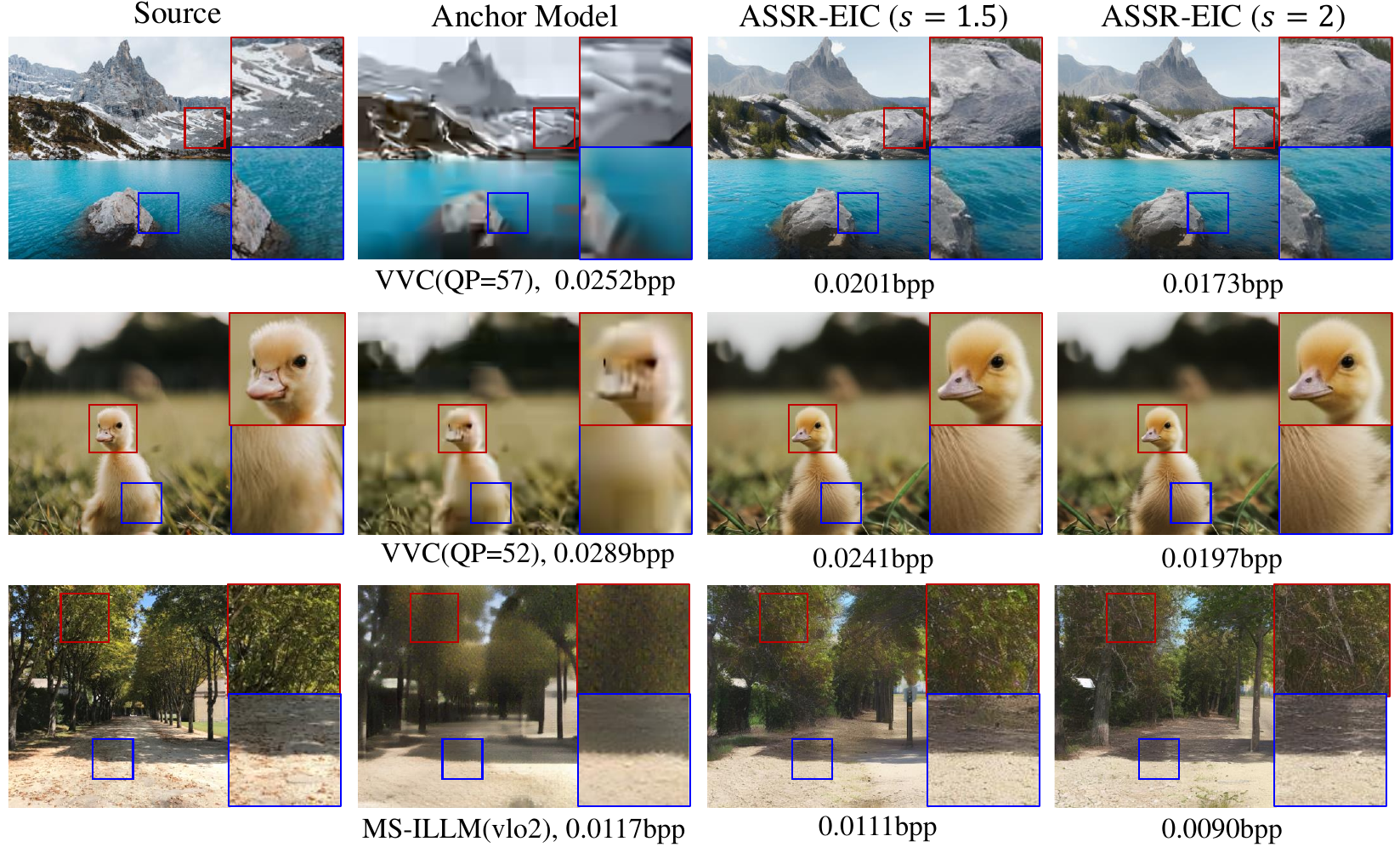}
   \caption{Subjective visual improvement results of ASSR-EIC against the basic anchor model under different rescaling factors $s$, showing its ability to maintain high perceptual realism even as bitrate decreases.}
   \label{fig:codec_subj_gain}
   \vspace{0.2em}
\end{figure*}

\begin{figure*}[htbp]
% \begin{minipage}[c]{0.96\linewidth}
  \centering
   \includegraphics[width=\linewidth]{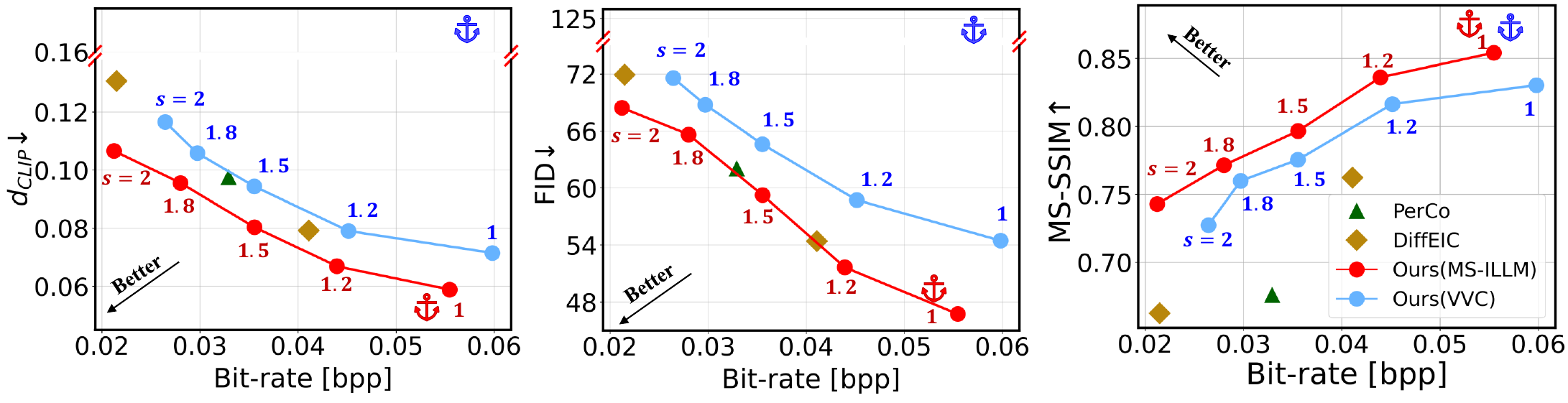}
   \caption{Impact of the rescaling factor $s$ on the bitrate and reconstruction quality of ASSR-EIC, evaluated using VVC and MS-ILLM as anchor codecs. 
    \raisebox{-0.2em}{\includegraphics[width=0.4cm]{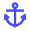}}~and~
    \raisebox{-0.2em}{\includegraphics[width=0.4cm]{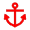}}
    represent the results of the VVC (QP=52) and MS-ILLM (“quality\_1”) anchor models, respectively.
    }
   \label{fig:control_obj}
   % \vspace{-1em}
% \end{minipage}
\end{figure*}

% \subsubsection{Training Protocol.}
% During training, we freeze all parameters of the pre-trained Stable Diffusion model and optimize only the newly introduced components: the image encoder, the auxiliary branch, the semantic attention modules, the RDS embedding networks, and the RDS modulation modules.

\section{Experiments}

\subsection{Experimental Settings}
\subsubsection{Selection of Anchor Methods} Our joint ASSR-enhanced image compression framework is compatible with any existing codec. In this work, we select a representative traditional codec, VVC \cite{JVET_VTM}, and a state-of-the-art learning-based regular-bitrate codec, MS-ILLM \cite{muckley2023improving}, to train a unified framework that supports both. 
% Notably, it requires no modification to the original encoder or decoder, enabling straightforward and fully compatible integration. 

\subsubsection{Implementation Details} We use BLIP-2 \cite{li2023blip} as the text encoder to extract image captions. For training data, we sample 100k images from the OpenImagesV6 dataset \cite{kuznetsova2020open} and randomly crop them to $512 \times 512$. Each cropped image is downsampled using bicubic interpolation with a continuous scale factor uniformly sampled from $[1, 2]$, and then degraded with VVC \cite{JVET_VTM} or MS-ILLM \cite{muckley2023improving} under diverse compression quality parameters to construct the training pairs.

We train ASSR-EIC using the Adam optimizer with a batch size of 12 and a learning rate of $5 \times 10^{-5}$. Training is conducted for 100k iterations on an NVIDIA A100 GPU, taking approximately two days. 
% We employ  to generate image captions and use SAM-Tiny \cite{kirillov2023segment} to extract semantic features. 

\subsubsection{Evaluation Details} We adopt three widely used benchmark datasets: Kodak \cite{Franzen1999}, CLIC2020 \cite{toderici2020workshop}, and MS-COCO \cite{lin2014microsoft}. For the CLIC2020 and MS-COCO datasets, due to the large variation in image resolutions, we follow a commonly adopted fixed-resolution evaluation protocol \cite{careil2024towards,li2024towards}. Specifically, we resize the shorter side to 512 and then apply center cropping to obtain 512×512 inputs. For the Kodak dataset, whose images are either 768×512 or 512×768, we perform full-resolution evaluation without resizing or cropping. To comprehensively assess both fidelity and perceptual quality, we report a set of commonly used metrics, including MS-SSIM \cite{wang2003multiscale}, LPIPS \cite{zhang2018unreasonable}, DISTS \cite{ding2020image}, CLIPScore $d_{\text{clip}}$ \cite{radford2021learning}, and FID \cite{heusel2017gans}. For $d_{\text{clip}}$, we compute the cosine similarity between CLIP image embeddings. 

\subsubsection{Compared Methods} 
We benchmark ASSR-EIC against both traditional codecs VVC \cite{JVET_VTM} and HEVC \cite{sullivan2012overview}, and state-of-the-art learning-based approaches ELIC \cite{he2022elic}, MS-ILLM \cite{muckley2023improving}, Text+Sketch \cite{lei2023text+}, PerCo \cite{careil2024towards}, and DiffEIC \cite{li2024towards}. Among these, Text+Sketch, PerCo, and DiffEIC are diffusion-based compression frameworks.
In the main analysis, we report results using MS-ILLM \cite{muckley2023improving} as the anchor codec.

\subsection{Quantitative Comparison}
We show quantitative comparison results on the Kodak, MSCOCO, and CLIC2020 datasets in Fig. \ref{fig:codec_obj}. Traditional non-generative codecs, such as VVC and HEVC, as well as learning-based methods like ELIC and MS-ILLM, exhibit a pronounced quality degradation at ultra-low bitrates, struggling to preserve high-realism content. In contrast, diffusion-based generative compression approaches deliver substantially better perceptual quality. Although Text+Sketch also incorporates diffusion priors, its reliance on sketch-only constraints leads to limited semantic and structural fidelity.

PerCo, DiffEIC, and our ASSR-EIC show clear advantages in this challenging regime. Across all datasets, our approach outperforms PerCo and DiffEIC, with notable gains in FID and CLIPScore. Moreover, unlike other diffusion-based codecs, our approach does not compromise pixel-level similarity in terms of MS-SSIM: it substantially exceeds PerCo and DiffEIC and remains comparable to pixel-level distortion-oriented codecs such as VVC, HEVC, and ELIC.

\subsection{Qualitative Comparison} 
We present visual comparison results with competing codecs in Fig. \ref{fig:codec_subj}. MS-ILLM, a state-of-the-art GAN-based codec, can reliably reconstruct coarse structures but often introduces blur and artifacts in fine details. Diffusion-based methods generally offer higher perceptual quality. However, Text+Sketch shows limited fidelity due to its reliance solely on sketch-based spatial constraints. PerCo and DiffEIC deliver stronger reconstructions, yet still fall short of our method in image fidelity, frequently exhibiting structural inconsistencies and color shifts. For instance, the chair structure in the first row is visibly distorted, and the starry sky in the second row appears noticeably desaturated in both PerCo and DiffEIC, deviating from the original image. 
In contrast, our method generates semantically coherent, structurally faithful, and detail-accurate reconstructions, achieving perceptually high-quality results at ultra-low bitrates while maintaining strong color consistency. One possible explanation is that, in these methods, lossy quantization is applied to diffusion latent representations. Since such latents are primarily optimized for generative modeling rather than strict signal fidelity, quantization perturbations in this space may influence the reconstruction characteristics and affect structure and color reconstruction behavior.

\subsection{Gains Compared with Anchor Codecs}
Tab. \ref{tab:gain} and Fig. \ref{fig:codec_obj_gain} demonstrate the performance gains achieved by our joint ASSR-enhanced image compression framework over the original anchor codecs. In this work, two representative anchors are used during training: the traditional codec VVC and the learning-based codec MS-ILLM. Tab. \ref{tab:gain} reports BD-rate results, where negative values indicate bitrate savings at equal perceptual quality. Fig. \ref{fig:codec_obj_gain} further visualizes the R–D improvements of our joint ASSR-enhanced framework over both VVC and MS-ILLM on the Kodak, MSCOCO, and CLIC2020 datasets.

\begin{table}[tbp]
    \centering
    \setlength{\tabcolsep}{10pt}
    \renewcommand{\arraystretch}{1.2}
    \caption{BD-Rate (\%) improvements achieved by our ASSR-EIC over the basic VVC and MS-ILLM anchor codecs on MSCOCO.}
    \begin{tabular}{cccc}
    \toprule
    \multirow{2}{*}{Anchor} & \multicolumn{3}{c}{ASSR-EIC (Anchor)}\\
    \cmidrule{2-4}
     & FID-BPP↓  & $d_{clip}$-BPP↓  & DISTS-BPP↓  \\
    \midrule
    VVC & -93.26 & -83.70 & -83.27 \\
    MS-ILLM & -30.96 & -31.90 & -14.44 \\
    \bottomrule
    \end{tabular}
    \label{tab:gain}
    \vspace{-1.5em}
\end{table}

As shown in Fig. \ref{fig:codec_obj_gain}, regardless of whether the anchor is a traditional or a learning-based codec, our method delivers substantial improvements in the ultra-low-bitrate regime. In particular, it reduces FID-Bitrate by 93.26\% and 30.96\% for VVC and MS-ILLM, respectively. Although the original VVC baseline performs significantly worse than learning-based methods, our framework elevates its performance to a level comparable with state-of-the-art approaches, highlighting the strong generalization and enhancement capability of our proposed ASSR-EIC.

Fig. \ref{fig:codec_subj_gain} illustrates the visual improvements achieved by enhancing the anchor codec with our ASSR-EIC framework. As shown, ASSR-EIC not only yields substantially higher subjective quality but also reduces the required bitrate.

It is also worth noting that, when the rescaling scale factor $s$ is set to 1, and bypasses the diffusion-based joint degradation-aware ASSR decoder (Section \ref{sec:rescaling}), the ASSR-EIC degenerates to the basic anchor codec. This preserves the anchor’s native encoding capability in the conventional bitrate range while keeping the entire process within a unified compression framework.

\subsection{Control Impacts of the Rescaling Factor $s$}
Our ASSR-EIC framework achieves variable-bitrate EIC through joint arbitrary-scale rescaling. The rescaling factor $s$ adjusts the bitrate by controlling the amount of spatial information discarded during encoding. 
Since $s$ can take any continuous float value, ASSR-EIC theoretically provides infinitely many operating points for achieving a desired bitrate. 
Fig. \ref{fig:control_obj} illustrates the influence of different $s$ values on ASSR-EIC. 
Our ASSR-EIC is compatible with arbitrary codec types and quality settings. In Fig. \ref{fig:control_obj}, representative compression-quality configurations of VVC and MS-ILLM are used as anchor models. By varying the value of $s$, we observe a clear trend: as 
$s$ increases, the bitrate consistently decreases while maintaining as much reconstruction quality as possible.

Fig. \ref{fig:codec_subj_gain} presents the subjective enhancement effects of ASSR-EIC applied to the anchor model under different values of $s$. As shown, even when $s$ increases, ASSR-EIC is still capable of producing high-quality reconstructions with strong perceptual realism. Nevertheless, a reduction in fidelity can be observed. For example, additional textures appear on the rocks on the water surface in the first row, and the duck’s feathers in the second row become over-sharpened compared with the original image. These results demonstrate that ASSR-EIC reconstructs images as faithfully as possible within the limited bitrate budget, while consistently preserving high perceptual realism.

% \begin{table}[t]
%     \centering
%     \setlength{\tabcolsep}{10pt}
%     \caption{Ablation studies on the core components on MS-COCO. We evaluate BD-Rate on the FID-BPP curve.}
%     \begin{tabular}{c|c}
%         \hline
%         Architecture & BD-Rate↓  \\ 
%         \hline
%         w/o Codec Embs in RDS Aligner & 2.77\% \\
%         w/o Rescaling Emb in RDS Aligner & 1.97\% \\
%         w/o RDS Modulator & 3.44\% \\
%         w/o Text Encoding &  6.48\% \\ 
%         w/o Semantic Attention & 6.69\%  \\ 
%         Full model &  \textbf{0\%} \\ \hline
%     \end{tabular}
%     \label{tab:abla}
%     % \vspace{-1em}
% \end{table}

\begin{table}[!t]
    \centering
    \renewcommand{\arraystretch}{1.2}
    \caption{Comparison with alternative super-resolution (SR) methods as the post-processing module after downsampling. We report BD-Rate (\%) on MS-COCO; lower is better.}
    \setlength{\tabcolsep}{4pt}
    \begin{tabular}{llccc}
    \toprule
    Anchor & SR method & FID-BPP$\downarrow$ & $d_{clip}$-BPP$\downarrow$ & DISTS-BPP$\downarrow$ \\
    \midrule
    \multirow{6}{*}{VVC} & -- (anchor only) & 0\% & 0\% & 0\% \\
     & RealESRGAN~\cite{wang2021real} & 7.74\% & -5.34\% & -9.81\% \\
     & PromptCIR~\cite{li2024promptcir} & 4.23\% & -0.68\% & 9.96\% \\
     & DA-CLIP~\cite{luo2024controlling} & -0.55\% & 0.67\% & 0.22\% \\
     & SeeSR~\cite{wu2024seesr} & -72.70\% & -55.49\% & -73.85\% \\
     & ASSR-EIC (Ours) & \textbf{-93.26\%} & \textbf{-83.70\%} & \textbf{-83.70\%} \\
    \midrule
    \multirow{6}{*}{MS-ILLM} & -- (anchor only) & 0\% & 0\% & 0\% \\
     & RealESRGAN~\cite{wang2021real} & 11.46\% & 18.40\% & 9.12\% \\
     & PromptCIR~\cite{li2024promptcir} & 1.91\% & 6.89\% & -4.74\% \\
     & DA-CLIP~\cite{luo2024controlling} & -0.41\% & -4.82\% & -2.80\% \\
     & SeeSR~\cite{wu2024seesr} & -15.17\% & 10.97\% & -13.56\% \\
     & ASSR-EIC (Ours) & \textbf{-30.96\%} & \textbf{-31.90\%} & \textbf{-14.44\%} \\
    \bottomrule
    \end{tabular}
    \label{tab:comp_SR}
    \vspace{0.2em}
\end{table}

\begin{table}[tbp]
    \centering
    \setlength{\tabcolsep}{10pt}
    \renewcommand{\arraystretch}{1.2}
    \caption{Generalization performance of our ASSR-EIC framework when applied to unseen codecs. BD-Rate (\%) results between our approach and anchor methods.}
    \begin{tabular}{cccc}
    \toprule
    \multirow{2}{*}{Anchor} & \multicolumn{3}{c}{ASSR-EIC (Anchor)}\\
    \cmidrule{2-4}
     & FID-BPP↓  & $d_{clip}$-BPP↓  & DISTS-BPP↓  \\
    \midrule
    HEVC & -87.97 & -76.14 & -87.29 \\
    ELIC & -89.20 & -82.66 & -86.92 \\
    \bottomrule
    \end{tabular}
    \label{tab:general}
    \vspace{1em}
\end{table}

\subsection{Distinguishing ASSR-EIC from Generic SR Post-Processing}
To clarify whether the observed performance gain originates from the proposed degradation-aware framework rather than from super-resolution (SR) alone, we compare ASSR-EIC with representative SR methods, including the GAN-based RealESRGAN \cite{wang2021real}, the Transformer-based PromptCIR \cite{li2024promptcir}, and diffusion-based approaches DA-CLIP \cite{luo2024controlling} and SeeSR \cite{wu2024seesr}. 
All methods are evaluated under the same downsampling–compression pipeline and operate on the decoded low-resolution images under identical bitrate settings.

As shown in Tab. \ref{tab:comp_SR}, generic SR methods yield limited or inconsistent BD-rate improvements across anchor codecs and evaluation metrics. Although SeeSR achieves strong perceptual gains when combined with VVC, a clear performance gap to ASSR-EIC remains. More importantly, these improvements are not consistently preserved: under MS-ILLM, the CLIPScore falls below the anchor-only baseline. 
These observations indicate that independently applied SR models, without access to encoding-stage information, lack robustness to codec-dependent compression characteristics. In contrast, ASSR-EIC integrates compression-related guidance into the reconstruction process through encoder–reconstruction coupling, enabling more stable and consistent bitrate reduction across anchor codecs.

\subsection{Generalization Across Other Anchor Methods}

To evaluate the generalization capability of our ASSR-EIC framework, we further applied the pre-trained model to compression methods that were not involved during training. As shown in Tab. \ref{tab:general}, the model trained only on VVC and MS-ILLM is directly applied to additional codecs, including the traditional HEVC codec and the learning-based ELIC codec, without any retraining. The results demonstrate consistent performance across these unseen anchor methods.

This generalization is enabled by the codec-agnostic reconstruction design of ASSR-EIC, complemented by a lightweight codec-aware conditioning mechanism. The proposed framework operates solely on the decoded outputs of the anchor codec, without relying on codec-specific latent representations. To accommodate the distinct artifact characteristics produced by traditional and learning-based codecs, we incorporate an explicit codec-type indicator ($\chi_{CT}$) as auxiliary conditioning. This minimal cue allows the diffusion-based reconstruction to adapt to different artifact distributions while preserving the codec-agnostic nature of the framework, thereby maintaining stable performance across diverse underlying anchor codecs.

\begin{table}[tbp]
    \centering
    \renewcommand{\arraystretch}{1.2}
    \caption{Ablation study of the dual semantic-enhanced design, showing the individual contributions of textual guidance and the soft semantic prompt, demonstrated by BD-Rate(\%).}
    \begin{tabular}{cccc}
    \toprule
    Method & FID-BPP↓  & $d_{clip}$-BPP↓  & DISTS-BPP↓  \\
    \midrule
    w/o Textual Guidance & 6.48 & 2.29 & 3.07 \\
    w/o Semantic Attention & 6.69 & 8.78 & 8.58 \\
    ASSR-EIC & \textbf{0} & \textbf{0} & \textbf{0} \\
    \bottomrule
    \end{tabular}
    \label{tab:ablation}
\end{table}

\begin{figure}[tbp]
  \centering
   \includegraphics[width=\linewidth]{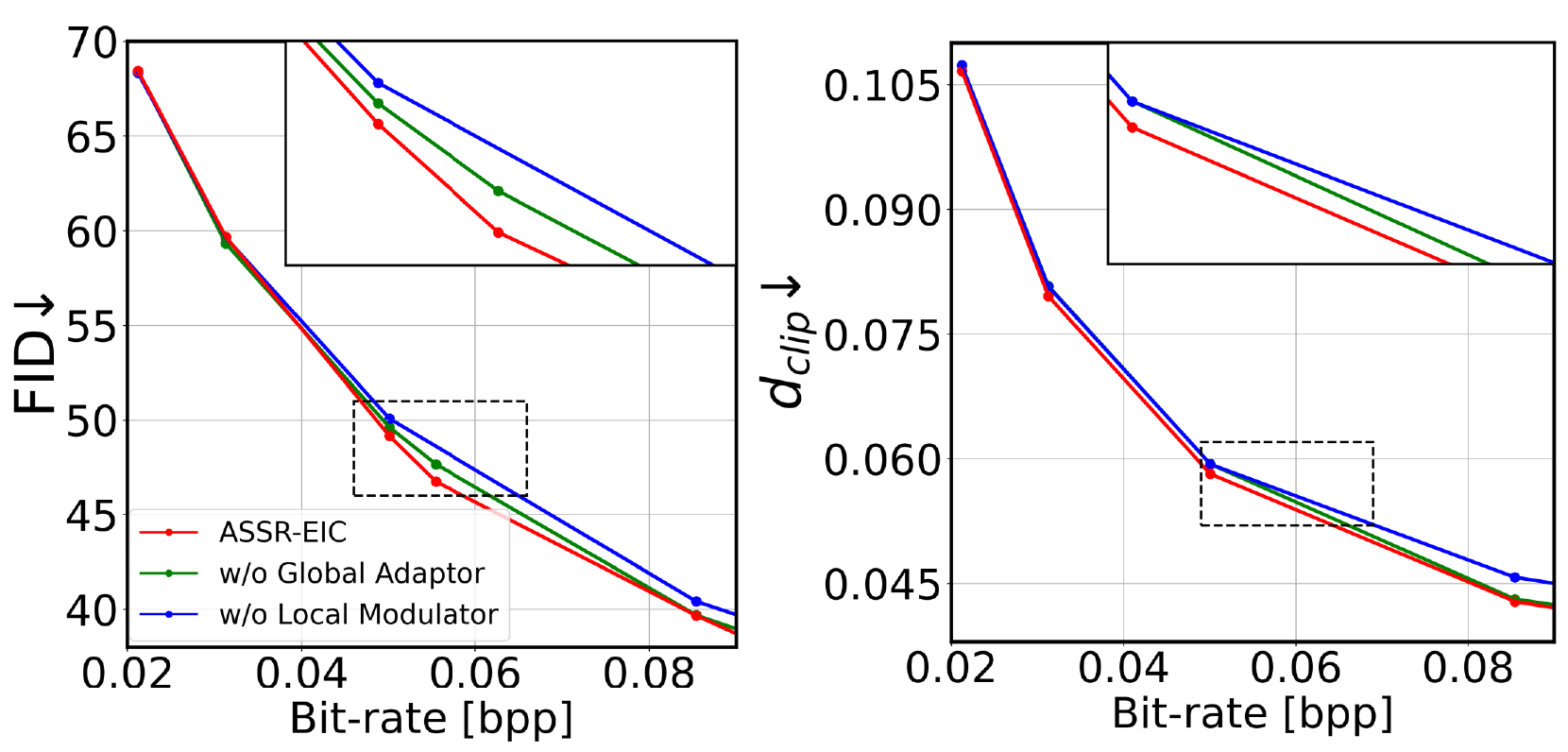}
   \caption{Ablation results of key components in ASSR-EIC, including the global compression–rescaling adaptor and the local compression–rescaling modulator, showing their impacts on FID and DISTS across different bitrates.}
   \label{fig:codec_ablation}
   % \vspace{-1em}
\end{figure}

\begin{table*}[htbp]
    \centering
    \caption{Complexity comparison between our ASSR-EIC framework and existing image compression methods, including training cost, encoding time, decoding time, and MACs. N denotes the number of models required to support different bitrates. The suffix “S” in diffusion-based methods (e.g., ASSR-EIC-S20 and ASSR-EIC-S50) denotes the number of diffusion inference steps.}
    \begin{tabular}{c|cccccc}
        \toprule
        Type & Method & Training Steps & Encoding Speed (s) & Decoding Speed (s) & Inference Step & MACs (G)  \\ 
        \midrule
        Non-Diffusion-based & VVC & - & 18.89±7.42 & 0.62±0.06 & 1 & - \\
        	& MS-ILLM & N×2M & 0.08±0.02 & 0.06±0.02 & 1 & 200 \\
            & ELIC & N×2M & 0.20±0.01 & 0.12±0.04 & 1 & 155 \\
        \midrule
        Diffusion-based & Text-Sketch & - & 137.28±2.55 & 9.45±0.53 & 20 & 303053 \\
            & PerCo & N×280k & 0.42±0.04	& 1.52±0.01 & 20 & 15934 \\
            & DiffEIC & N×500k & 0.33±0.01 & 3.05±0.02 & 50 & 36491 \\
            & ASSR-EIC-S20 (Ours) & 100k	& 0.44±0.06	& 3.41±0.16 & 20 & 27255 \\
            & ASSR-EIC-S50 (Ours) & 100k	& 0.44±0.06	& 6.96±0.41 & 50 & 65634 \\
        \bottomrule
            
        % w/o Codec Embs in RDS Aligner & 2.77\% \\
        % w/o Rescaling Emb in RDS Aligner & 1.97\% \\
        % w/o RDS Modulator & 3.44\% \\
        % w/o Text Encoding &  6.48\% \\ 
        % w/o Semantic Attention & 6.69\%  \\ 
        % Full model &  \textbf{0\%} \\ \hline
    \end{tabular}
    \label{tab:complex}
    % \vspace{-1em}
\end{table*}

\subsection{Ablation Studies}

\subsubsection{Effectiveness of the global compression-rescaling adaptor} 
The global compression-rescaling adaptor is designed to provide global compression-aware and rescaling-aware guidance during the diffusion-based generation process. To verify its effectiveness, we replace it with the original ResNet module from Stable Diffusion. Fig. \ref{fig:codec_ablation} shows the results on MSCOCO, this substitution led to noticeable degradation in both FID and DISTS, particularly in the bitrate range above 0.04 bpp. As shown, the global compression-rescaling adaptor significantly improves fidelity and distribution similarity, demonstrating its crucial role in our framework.

\subsubsection{Effectiveness of the local compression-rescaling modulator} 
The local compression–rescaling modulator is designed to dynamically adjust the diffusion process according to the image’s encoding and rescaling conditions, working in conjunction with the main generation branch and the fidelity module to enhance detail reconstruction. To evaluate its effectiveness, we removed this modulator and present the results in Fig. \ref{fig:codec_ablation}. As shown, the absence of the local compression–rescaling modulator leads to a notable performance drop, whereas its inclusion yields substantial improvements in both FID and DISTS, particularly in the range above 0.03 bpp.

\subsubsection{Effectiveness of the dual semantic-enhanced design}
Our dual semantic-enhanced design aims to improve reconstruction fidelity. It consists of two components: a textual guidance module that leverages a image caption as the text prompt, and a soft semantic prompt that injects semantic attention derived from SAM image-encoder features. To evaluate the contribution of each component, we removed them individually, with the results on MSCOCO summarized in Tab. \ref{tab:ablation}. As shown, both forms of guidance play critical roles in enhancing fidelity. In particular, the soft semantic prompt provides substantial improvements in FID, CLIPScore, and DISTS, demonstrating its strong effectiveness in preserving semantic structure and perceptual quality.

\subsection{Complexity Analysis}
Tab.~\ref{tab:complex} summarizes the complexity comparison between our method and existing image compression approaches. A key advantage of our framework is its ability to support variable-bitrate compression within a single unified model. This design removes the need to train multiple diffusion models for different bitrate levels and leads to a substantial reduction in overall training cost compared with prior diffusion-based EIC methods.

In terms of encoding time, the reported results include textual encoding, anchor codec encoding, and spatial downsampling. Under this complete configuration, our approach achieves encoding efficiency comparable to recent diffusion-based EIC methods. When textual guidance is disabled by providing an empty prompt, the encoding time closely matches that of the underlying anchor codec because spatial downsampling introduces negligible overhead. In this configuration, the encoding time is 0.08 s when using MS-ILLM as the anchor codec, which is lower than that of recent diffusion-based approaches. This comes at the cost of moderate performance degradation of approximately 2.3\%–6.5\%, as shown in the first row of Tab. \ref{tab:ablation}.

Regarding decoding, our method maintains a reasonable runtime given its ability to deliver high-quality, scale-adaptive reconstruction. The current implementation adopts 50 diffusion steps to ensure strong perceptual performance, which increases computation. However, the decoding process can be further accelerated. Recent progress in diffusion acceleration, including fast sampling techniques and diffusion model distillation, offers practical means to reduce inference time. Integrating these advances is a promising direction for enhancing the efficiency of our framework in future work.

% \begin{figure}[t]
%   \centering
%    \includegraphics[width=\linewidth]{imgs/infer_controlV2.pdf}

%    \caption{Benefiting from our continuous decoding representation, our rescaling diffusion decoder enables adjustable enhancement with more realistic details by applying more extreme paramters during decoding process.}
%    \label{fig:infer_control}
%    % \vspace{-0.5em}
% \end{figure}

\section{Conclusion and Limitations}
In this paper, we introduce arbitrary-scale image rescaling into image compression and propose a novel variable-bitrate EIC framework that is compatible with both traditional and learning-based codecs. On the encoder side, an arbitrary-rate spatial downsampling module is incorporated to achieve controllable bitrate reduction. On the decoder side, we develop a diffusion-based, joint degradation-aware ASSR reconstruction decoder that enables high-quality, original-resolution reconstruction with effective rate adaptation. To further enhance reconstruction fidelity, we integrate a dedicated fidelity module and adopt a dual semantic-enhanced design. In addition, to improve bitrate adaptivity, we propose a global compression–rescaling adaptor and a local compression–rescaling modulator, which together facilitate flexible and robust rate-adaptive reconstruction. Experimental results demonstrate that our method achieves state-of-the-art ultra-low-bitrate compression while providing flexible bitrate scalability across diverse operating conditions.

The main limitation of our method lies in its relatively long decoding time, which is primarily attributed to the large number of inference steps required by the diffusion-based model. Incorporating recent advances in accelerating diffusion models, such as single-step sampling and quantization techniques, offers a promising direction for further improving efficiency in our future work.

% \section*{Acknowledgments}
% This should be a simple paragraph before the References to thank those individuals and institutions who have supported your work on this article.

% {\appendix[Proof of the Zonklar Equations]
% Use $\backslash${\tt{appendix}} if you have a single appendix:
% Do not use $\backslash${\tt{section}} anymore after $\backslash${\tt{appendix}}, only $\backslash${\tt{section*}}.
% If you have multiple appendixes use $\backslash${\tt{appendices}} then use $\backslash${\tt{section}} to start each appendix.
% You must declare a $\backslash${\tt{section}} before using any $\backslash${\tt{subsection}} or using $\backslash${\tt{label}} ($\backslash${\tt{appendices}} by itself
%  starts a section numbered zero.)}


\bibliographystyle{IEEEtran}
\bibliography{reference}

\end{document}